\documentclass[runningheads]{llncs}

\usepackage{eccv}

\usepackage{eccvabbrv}

\usepackage{graphicx}
\usepackage{booktabs}

\usepackage[accsupp]{axessibility}  

\usepackage{hyperref}

\usepackage{orcidlink}

\definecolor{purple}{rgb}{0.48,0.14,0.98}
\definecolor{dgreen}{rgb}{0.06,0.51,0.03}

\definecolor{SigmaColor}{rgb}{0.98,0.45,0.0}

\newcommand{\parA}[1]{\left(#1\right)}
\newcommand{\parB}[1]{\left\{#1\right\}}

\newcommand{\myvspace}[1]{\vspace{#1}}
\renewcommand{\paragraph}[1]{\noindent\textbf{#1}}
\usepackage[misc]{ifsym}
\def \olt {\mathcal{OLT}}

\def \query {\mathcal{Q}}
 
\def \fvlm {\mathbf{VLM}}
\def \fseg {\mathbf{SEG}}

\def \bbox {\mathbf{bbox}}
\def \pcd {\mathcal{P}}
\def \rgbset {\mathcal{I}}
\def \rgb {\mathbf{I}}
\def \maskset {\mathcal{M}}

\def \id {\mathbf{ID}}
\def \objlabel {\mathbf{label}}
\def \objidset {\mathcal{C}}
\def \taskchain {\mathcal{T}}
\usepackage{booktabs} 
\usepackage{multirow}
\usepackage{makecell}
\usepackage{ulem}
\usepackage{utfsym}
\usepackage{adjustbox}
\usepackage{tabularx}
\usepackage{algorithm}
\usepackage{algpseudocode}

\usepackage{longtable}
\usepackage[most]{tcolorbox}
\usepackage[absolute,overlay]{textpos}
\usepackage{xcolor}

\definecolor{headerbg}{RGB}{70,70,70}
\definecolor{headerfg}{RGB}{255,255,255}
\definecolor{boxbg}{RGB}{245,245,245}

\DeclareCaptionType{myempty}[][]
\begin{document}

\title{OpenGround: Planning-based Online Perception for Open-World 3D Visual Grounding} 

\titlerunning{OpenGround}

\author{Wenyuan Huang\inst{1} \and
Zhenyu Zhang\inst{1*} \and
Zhao Wang\inst{2} \and
Zhou Wei\inst{2} \and
Ting Huang\inst{1} \and
Fang Zhao\inst{1} \and
Jian Yang\inst{1}}

\authorrunning{W. Huang et al.}

\institute{Nanjing University, School of Intelligent Science and Technology, Nanjing, China \and
China Mobile Zijin Innovation Institute, Nanjing, China\\
*Corresponding Author\\
\email{wenyuan\_huang@smail.nju.edu.cn}, \email{zhenyuzhang@nju.edu.cn}, \email{wangzh8@js.chinamobile.com}, \email{zhouwei9@js.chinamobile.com}, \email{huangting0403@sues.edu.cn}, \email{zhaofang0627@gmail.com}, \email{csjyang@nju.edu.cn} \\
Project Page: \href{https://why-102.github.io/openground.io/}{\textcolor[HTML]{CB3189}{https://why-102.github.io/openground.io/}}
}

\maketitle

\begin{abstract}
3D visual grounding aims to locate objects based on natural language descriptions in 3D scenes. Existing supervised methods are limited by generalization and recent zero-shot methods typically rely on a predefined Object Lookup Table (OLT) to query Visual Language Models (VLMs) for reasoning about object locations via a single step grounding, which limits the applications in scenarios with undefined targets and complex queries.
To address these problems, we present OpenGround, a novel zero-shot framework for open-world 3D visual grounding that remains compatible with recent zero-shot methods. OpenGround integrates Task-Chain Planning to decompose a query into a plan of context-to-target sub-goals for progressive grounding, and Context-Guided Perception to perceive novel objects online under context guidance from the task chain.
We also propose a new dataset named OpenTarget, which contains over 7000 object-description pairs to mimic open-world evaluation. 
Extensive experiments demonstrate that OpenGround achieves competitive performance on Nr3D, state-of-the-art on ScanRefer, and delivers a substantial 17.6\% improvement on OpenTarget.
\keywords{3D Visual Grounding \and Vision-Language Model \and Zero-shot Scene Understanding}
\end{abstract}
\begin{figure}[t]
  \centering
  \includegraphics[width=\linewidth]{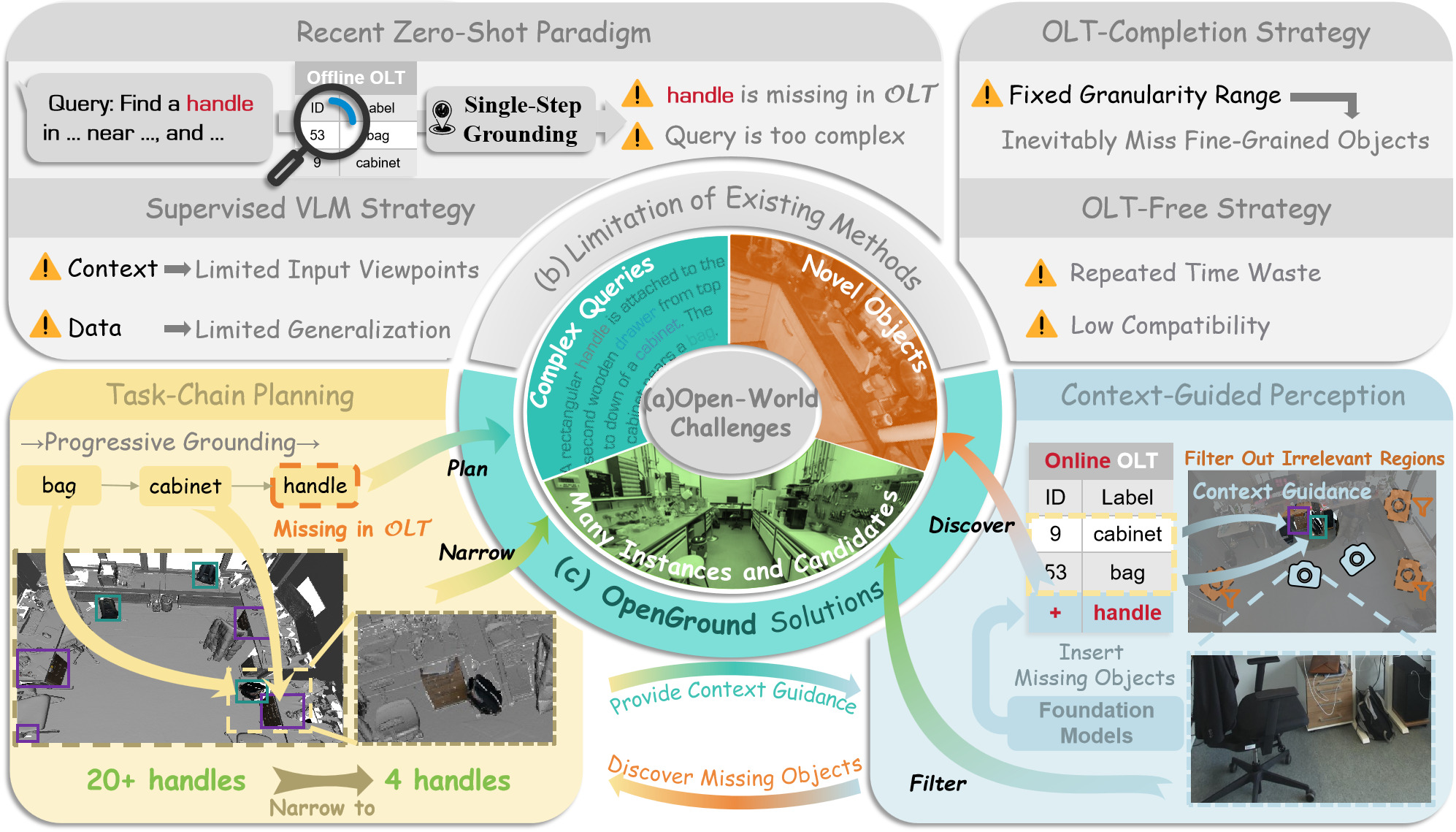} 
    \caption{\textbf{Overview of Introduction.} (a) Open-world 3D visual grounding faces challenges from complex queries, novel objects, and massive instances and candidates. (b) Existing methods are limited: supervised 3D-VLMs suffer from data scarcity and restricted input views; recent zero-shot pipelines rely on an offline OLT and break when the target is missing or the query is compositional, and either OLT-Completion Strategy is impractical or OLT-Free Strategy repeatedly scans the whole scene and is less compatible with OLT-based backbones.  (c) Our proposed OpenGround framework addresses these limitations through Task-Chain Planning, which progressively grounds sub-goals to narrow the search space, and Context-Guided Perception, which dynamically discovers missing objects and turns a static offline OLT into an online OLT.}
  \label{fig:teaser}
\end{figure}

\section{Introduction}
\label{sec:intro}

3D Visual Grounding (3DVG) aims to locate target objects in 3D scenes from natural language queries, with broad applications in AR/VR \cite{10553387, unal2024ways, chen2020scanrefer, ma2023examination}, vision-language navigation \cite{zhang2024vision, huang2022assister, huang2022visual} and robotic perception \cite{chen2023clip2scene, kong2023robo3d, liu2025aligning}. 
A practical 3DVG approach should not only reason over natural language queries but also flexibly adapt to targets from coarse object categories to fine-grained parts, and generalize to previously unseen real world environments.

Research in 3DVG has made significant progress, encompassing both supervised and zero-shot methods. 
While supervised methods \cite{huang20253d, huang2025viewsrd, licityanchor, lin2025groundflow, zheng2025densegrounding, zhu2025move, qi2025gpt4scene, guo2025text}, trained on diverse datasets \cite{dai2017scannet, chen2020scanrefer, achlioptas2020referit3d, zhang2023multi3drefer, solmaz2025scanverse, baruch2021arkitscenes}, and zero-shot methods \cite{li2025seeground, lin2025seqvlm, zhu2025struct2d, zantout2025sort3d, shi2025chain, liu2025reasongrounder, xu2024vlmgrounder, wang2025affordbot} leveraging the power of VLMs, both achieve high accuracy on existing benchmarks.
However, in practical deployment, they are facing open-world challenges: as illustrated in \cref{fig:teaser}-(a), there are long and complex queries, novel and fine-grained objects, and massive instances and candidates.

Under such challenges, supervised methods face two practical constraints, in \cref{fig:teaser}-(b) ``Supervised VLM Strategy'': first, supervision scarcity limits real-world generalization; second, context window of supervised 3D-VLM limits its number of input views, probably missing critical evidence in large scenes.
Further, recent zero-shot methods \cite{li2025seeground, lin2025seqvlm, zhu2025struct2d, zantout2025sort3d, shi2025chain, liu2025reasongrounder} adopt a paradigm in \cref{fig:teaser}-(b) ``Recent Zero-Shot Paradigm'': they use 2D VLMs \cite{openai2024gpt4technicalreport, Qwen2-VL, Qwen2.5-VL} for single-step grounding with a predefined object lookup table ($\olt$). 
However, they are fundamentally limited by the predefined offline OLT which is built on a fixed granularity range before queries and inevitably misses novel and fine-grained objects, even though a stronger segmentor is provided by AffordBot \cite{wang2025affordbot} in \textbf{OLT-Completion Strategy}. At the other extreme, \textbf{OLT-Free Strategy}, such as VLM-Grounder \cite{xu2024vlmgrounder}, abandons $\olt$ entirely but suffers from repeated searching whole scene for common objects and less compatible with recent OLT-based methods. Moreover, single-step formulation is brittle for complex queries which require multi-step evidence gathering.
These suggest an intuitive solution: instead of exhaustively completing $\olt$ or abandoning it in a single-step grounding, a practical open-world 3DVG system should plan and ground progressively to resolve complex queries, and discover novel objects to extend $\olt$ online.

In this paper,we propose OpenGround, a zero-shot framework designed to address the open-world challenges of 3D visual grounding through two key modules, while remaining compatible with existing OLT-based pipelines.
As illustrated in \cref{fig:teaser}-(c), given a query, Task-Chain Planning decomposes it into a plan with a sequence of sub-goals. By grounding through the plan progressively,  OpenGround narrows the search space for the final target.
Complementing the task chain, Context-Guided Perception (CGP) discovers missing objects by leveraging previously grounded objects as context guidance. When a sub-goal does not exist in $\olt$, CGP filters out irrelevant regions to perceive the missing object and inserts it into the table online. It effectively transforms a static offline OLT into a dynamic online OLT.
As a result, subsequent single-step grounding can continue on the updated online $\olt$, enabling efficient grounding of both predefined and previously unseen objects under complex open-world queries.

To evaluate our framework, we also construct a novel benchmark named OpenTarget to simulate open-world evaluation. The benchmark takes ScanNet++ \cite{yeshwanthliu2023scannetpp} as the basement, and integrate fine-grained part-level segmentation results from Articulate3D \cite{halacheva2024articulate3d} as the targeted but undefined 3D objects to simulate the open-world setting. With the automatic description generating and rigorous quality filtering, we obtain 7,724 query-object pairs which are sufficiently comprehensive to support reliable validation experiments. Overall, our contributions are summarized as follows:
\begin{itemize}
    \item We define a novel task for zero-shot 3DVG: grounding objects outside the predefined object lookup table ($\olt$) in 3D environments using natural language descriptions. We further propose a benchmark named OpenTarget to enable rigorous evaluation of the open-world 3DVG task. The dataset uses VLM-generated natural language descriptions, and ensures final quality through manual filtering, resulting in 7,724 high-quality query-object pairs.
    \item We extend the existing zero-shot 3DVG paradigm by integrating the Context-Guided Perception (CGP), which enables grounding objects beyond $\olt$.
    \item We propose Task-Chain Planning for progressive grounding to resolve complex queries. It also restricts CGP application to regions around already grounded context objects, balancing efficiency and precision.
\end{itemize}
\section{Related Work}
\label{sec:relwork}

\paragraph{Supervised 3DVG.}
Supervised 3D visual grounding began with ScanRefer \cite{chen2020scanrefer} and Referit3D \cite{achlioptas2020referit3d}, and can be categorized into two paradigms: two-stage and single-stage methods. Two-stage approaches \cite{chen2020scanrefer, achlioptas2020referit3d, jain2022bottom, licityanchor, zhao20213dvg, zhang2023multi3drefer, zhu20233dvista} first generate object proposals via 3D segmentation \cite{liu2021group, vu2022softgroup, jiang2020pointgroup, schult2022mask3d}, then match them to textual queries. In contrast, single-stage methods \cite{qian2024multi, wang2024g, lin2025groundflow, zheng2025densegrounding, huang2025viewsrd, qi2025gpt4scene, guo2025text, huang20253d, unal2024four} employ end-to-end architectures that jointly learn 3D and language representations for direct grounding. Recent advances \cite{chen2025gsreasoner,huang20253d, qi2025gpt4scene, zhu2024llava, xu2024pointllm} further integrate VLMs to enhance 3D understanding, forming 3D-VLMs. 
Notably, GS-Reasoner \cite{chen2025gsreasoner} leverages chain-of-thought reasoning to better handle long and complex queries; however, such 3D-VLM solutions limit number of input views and involve expensive, hard-to-control CoT inference, which restricts evidence gathering and practical deployment. Moreover, despite their strong performance, particularly 3D-R1 \cite{huang20253d} which achieves SoTA through large-scale GRPO \cite{deepseek-math} training, these methods remain constrained by training data and fail to generalize in open-world grounding.

\paragraph{Zero-Shot 3DVG.}
Zero-shot 3DVG aims to improve generalization beyond supervised 3DVG by leveraging pre-trained VLMs. Early attempts \cite{Peng2023OpenScene, conceptfusion} follow CLIP-based paradigms, where query is matched against 3D proposals or fused scene features via cross-modal similarity, enabling zero-shot grounding but suffering from multiple candidates.
Recent trend \cite{li2025seeground, lin2025seqvlm, zhu2025struct2d, zantout2025sort3d, shi2025chain, liu2025reasongrounder} uses an \textbf{offline object lookup table ($\olt$)} to assist VLMs: methods first retrieve candidates from the $\olt$ (from 3D segmentation or CLIP-based detection), then perform VLM-based single-step confirmation to ground the target, achieving strong accuracy and improved efficiency in closed-set settings.
However, the reliance on a predefined $\olt$ fundamentally limits open-world deployment, since undefined or long-tailed objects (especially fine-grained parts) are not covered.
Existing remedies fall into two extremes: AffordBot \cite{wang2025affordbot} densifies $\olt$ with a stronger segmentor but still inevitably misses novel objects; VLM-Grounder \cite{xu2024vlmgrounder} abandons $\olt$ at the cost of repeated full-scene scanning and lower compatibility with modular OLT-based pipelines.
Moreover, they remain single-step and struggle with complex queries, which is common in open-world deployment.

\paragraph{Advances in VLMs and Visual Perception.}
Recent VLMs \cite{vteam2025glm45v,liu2023llava, liu2023improvedllava, liu2024llavanext,Qwen-VL, Qwen2-VL, Qwen2.5-VL, qwen3technicalreport,chen2024internvl, wang2025internvl3_5,step3system}, such as the Qwen-VL series, show strong cross-domain generalization and fine-grained spatial understanding \cite{qwen3technicalreport, wang2025internvl3_5}. Meanwhile, 3D perception frameworks like PointGroup \cite{jiang2020pointgroup}, Mask3D \cite{schult2022mask3d}, and Point-SAM \cite{zhou2025pointsam} have advanced but still lack open-world generalization due to limited 3D data. In contrast, 2D foundation models \cite{kirillov2023segany, ravi2024sam2, huang2025deimv2, liu2023grounding, ren2024grounded}, such as SAM \cite{kirillov2023segany, ravi2024sam2}, achieve strong open-world segmentation and grounding. The convergence of spatially aware VLMs and generalizable 2D perception models makes open-world 3D grounding promising.

\section{Dataset}
\label{sec:dataset}

To evaluate the open-world grounding capability of OpenGround, we construct a novel dataset named OpenTarget to \textbf{mimic} open-world settings, based on ScanNet++ \cite{yeshwanthliu2023scannetpp} and Articulate3D \cite{halacheva2024articulate3d}. Existing 3DVG benchmarks \cite{chen2020scanrefer, achlioptas2020referit3d} focus on object-level instances with limited category diversity, failing to simulate the open-world scenarios where fine-grained, undefined objects (\eg, sink handles, cabinet doors) are pervasive. In contrast, OpenTarget introduces objects from Articulate3D absent in the $\olt$ built on ScanNet++. These fine-grained parts \textbf{mimic} unforeseen objects in open-world scenarios, providing a benchmark for open-world grounding. OpenTarget provides totally 7,724 object-pairs for segments in Articulate3D, across 50 object classes and 70 part classes.
We outline the dataset construction and partial statistics below and details in \textbf{Appendix}.

\begin{figure}[t]
  \centering
  \includegraphics[width=\linewidth]{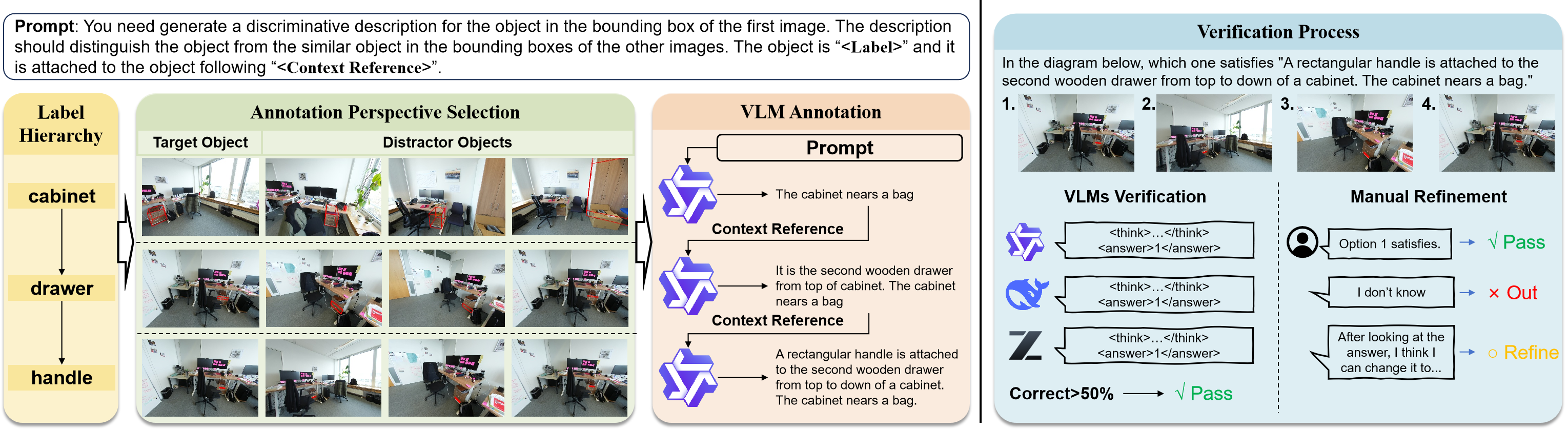}
  \includegraphics[width=\linewidth]{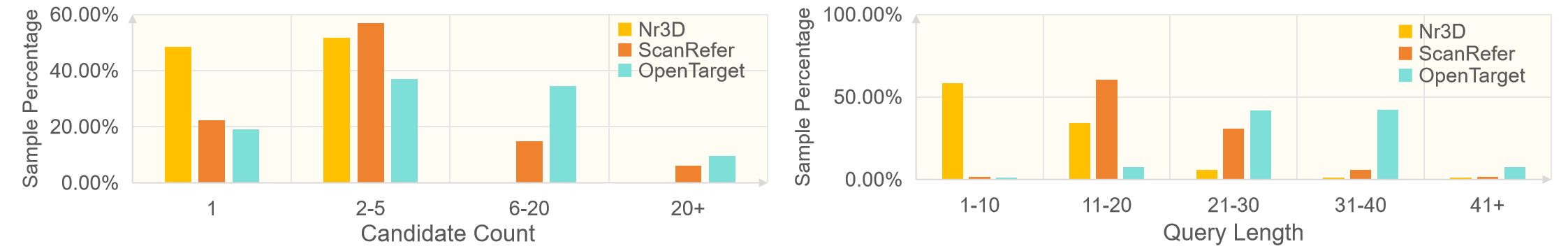}
  \caption{\textbf{OpenTarget dataset construction and statistics.}
\textbf{Top:} A progressive VLM-assisted pipeline builds discriminative queries for part-level targets from Articulate3D on ScanNet++ scenes, using a label hierarchy (\eg, cabinet$\rightarrow$drawer$\rightarrow$handle), target–distractor viewpoint selection, and context-aware description generation conditioned on parent annotations, followed by VLM voting and human refinement.
\textbf{Bottom:} Compared with Nr3D and ScanRefer, OpenTarget shows larger candidate pools and longer, more compositional queries, reflecting harder grounding.}
  \label{fig:data_collection_pipeline}
\end{figure}

\paragraph{Annotation.} As illustrated in \cref{fig:data_collection_pipeline}, we adopt a progressive VLM-based strategy for annotation, involving three stages: Label Hierarchy, Annotation Perspective Selection, and VLM Annotation. First, we utilize the hierarchical labels from Articulate3D (\eg, cabinet$\rightarrow$drawer$\rightarrow$handle), starting from top-level objects in subsequent processes. Second, we collect the target and its distractors, and select suitable perspectives for each. Third, we prompt VLMs with these perspectives and labels to generate queries that distinguish the target from distractors. For child objects, we repeat the process with parent object annotations as contextual references, ensuring progressively detailed and discriminative queries.

\paragraph{Verification.} We employ a two-stage quality verification process, combining automatic filtering and manual refinement. First, for objects with the same label, we use multiple VLMs to vote on object-query pairs (via selected perspectives and object IDs), retaining only majority-approved pairs to manual review. Subsequently, human annotators verify these pairs, with the ability to mark them as ``unidentifiable'' or revise queries, ensuring high dataset quality.

\paragraph{Statistics.}
OpenTarget contains 7,724 object-query pairs spanning 50 object classes and 70 part classes, built on ScanNet++ scenes with part-level objects from Articulate3D.
Compared with object-level 3DVG benchmarks, OpenTarget poses a challenging open-world setting, since targets are fine-grained parts that are absent from the $\olt$ and the training datasets of 3D segmentor, and appear higher ambiguity.
As shown in \cref{fig:data_collection_pipeline}, OpenTarget exhibits a larger number of same-label instances per scene (average \textbf{8.35} vs. \textbf{4.13} on ScanRefer and \textbf{3.03} on Nr3D), which substantially increases distractors for grounding.
Meanwhile, OpenTarget queries are longer (median length \textbf{31} words vs. \textbf{19} on ScanRefer and \textbf{9} on Nr3D), reflecting possible multi-step reasoning requirement.
These statistics highlight that OpenTarget evaluates open-world grounding under both higher visual ambiguity and more complex queries.

\section{Preliminary}
\label{sec:preliminary}

\begin{figure}
  \centering
  \includegraphics[width=\linewidth]{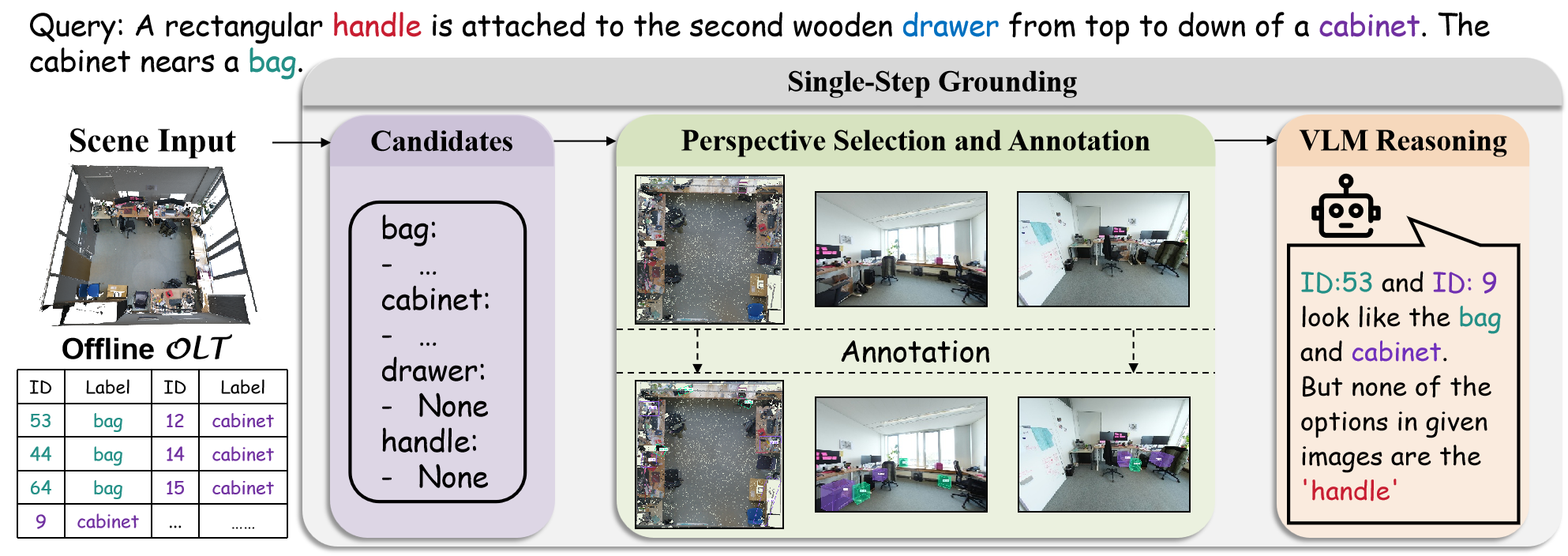}
  \myvspace{-8pt}
  \caption{\textbf{Single-Step Grounding.} Recent zero-shot methods \cite{li2025seeground, lin2025seqvlm, zhu2025struct2d, zantout2025sort3d, shi2025chain, liu2025reasongrounder} follow the paradigm: decompose the query into objects and retrieve candidates from the predefined $\olt$, then select informative views to observe and annotate candidates, finally find target ID through VLM reasoning and obtain 3D bounding box from $\olt$.}
  \label{fig:single-step-grounding}
  \myvspace{-16pt}
\end{figure}

OpenGround is built upon the OLT-based Single-Step Grounding paradigm which is widely adopted in many recent zero-shot 3DVG methods, and extends it to handle open-world targets and complex queries.
Given a 3D scene $S$ (\eg, point cloud) and a natural-language query $\query$, they first construct an object lookup table ($\olt$) via existing 3D scene segmentation model, denoted as 
\begin{equation}
    \olt=\parB{(\id_i,\objlabel_i,\bbox_i)}_{i=1}^N,
\end{equation}
which maps object IDs to their semantic labels and 3D bounding boxes.

As illustrated in \cref{fig:single-step-grounding}, the paradigm consists of three stages.

\paragraph{Candidate Retrieval.} The paradigm first extracts objects mentioned in $\query$ and decomposes them into anchors $\mathcal{A}$ and targets $\mathcal{T}$, where anchors provide references and targets denote the objects to be grounded (\eg, in ``find the \textit{laptop} in front of the \textit{chair}'', $\textit{chair}\in\mathcal{A}$ and $\textit{laptop}\in\mathcal{T}$). Then, it retrieves candidates of these objects from $\olt$ and filters out impossible candidates.

\paragraph{Perspective Selection and Annotation.} Given retrieved candidates, the paradigm selects informative views that reveal these candidates, so that VLM can distinguish them. The selected views are rendered from the 3D scene, and each candidate’s 3D bounding box is projected onto the corresponding images using camera poses, producing 2D visual annotations (\eg, boxes with instance IDs). The annotated multi-view images explicitly indicate the coordinates of each candidate, serving as the visual evidence for the subsequent VLM reasoning.

\paragraph{VLM Reasoning.}
Finally, the paradigm feeds $\query$ and the annotated multi-view images into a VLM, which performs single-step grounding by selecting the most plausible instance ID among the annotated candidates according to the semantics and spatial relations. The predicted ID is then mapped back to its corresponding 3D bounding box in $\olt$, yielding the target in the 3D scene.

\section{Methodology}
\begin{figure}
  \centering
  \includegraphics[width=\linewidth]{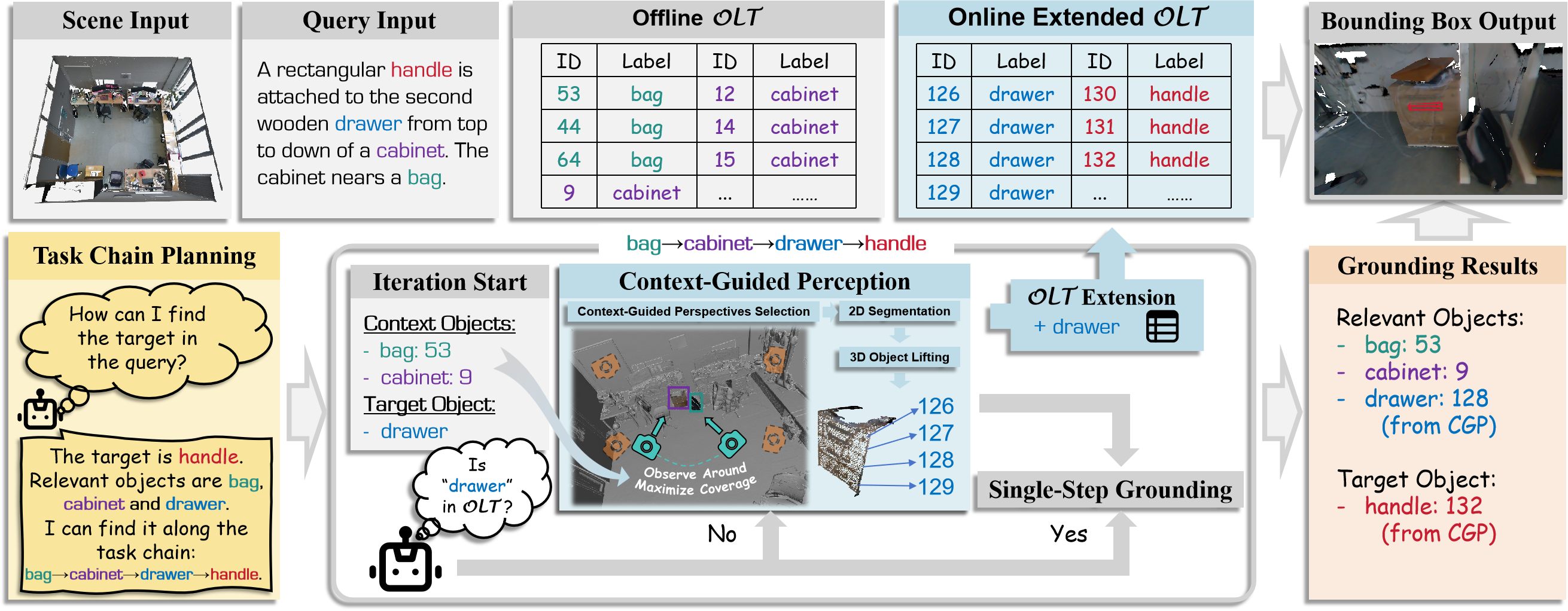}
  \myvspace{-8pt}
  \caption{\textbf{Overview of OpenGround.} OpenGround keeps the OLT-based single-step grounding backbone and augments it with Task-Chain Planning and Context-Guided Perception (CGP).
  It first plans an ordered sequence of sub-goals, then iterates over the sequence.
  If a sub-goal is missing from the current $\olt$, CGP discovers new objects around grounded objects and extends the $\olt$ online, after which single-step grounding proceeds on the extended $\olt$ to retrieve the 3D bounding box.}
  \label{fig:overview}
  \myvspace{-16pt}
\end{figure}

\paragraph{Overview.}
OpenGround builds on the OLT-based single-step grounding in \cref{sec:preliminary} and extends it to open-world 3DVG with two components in \cref{fig:overview}:
\textbf{Task-Chain Planning} plans an ordered sequence of sub-goals over context objects and the final target, so grounding can proceed step-by-step and each sub-goal continuously narrows the remaining search space for subsequent steps; 
\textbf{Context-Guided Perception (CGP)} discovers novel objects under context guidance from task chain, which moves the offline OLT to an online OLT. When the current sub-goal does not exist in $\olt$, CGP leverages already grounded context objects to filter out irrelevant regions and performs perception to discover missing objects and extend the $\olt$ on-the-fly. 
Subsequent single-step grounding then continues on the extended online $\olt$, enabling grounding of both predefined and previously unseen objects under complex open-world queries.

\begin{figure}
  \centering
  \begin{subfigure}[t]{0.49\linewidth}
    \centering
    \includegraphics[width=\linewidth]{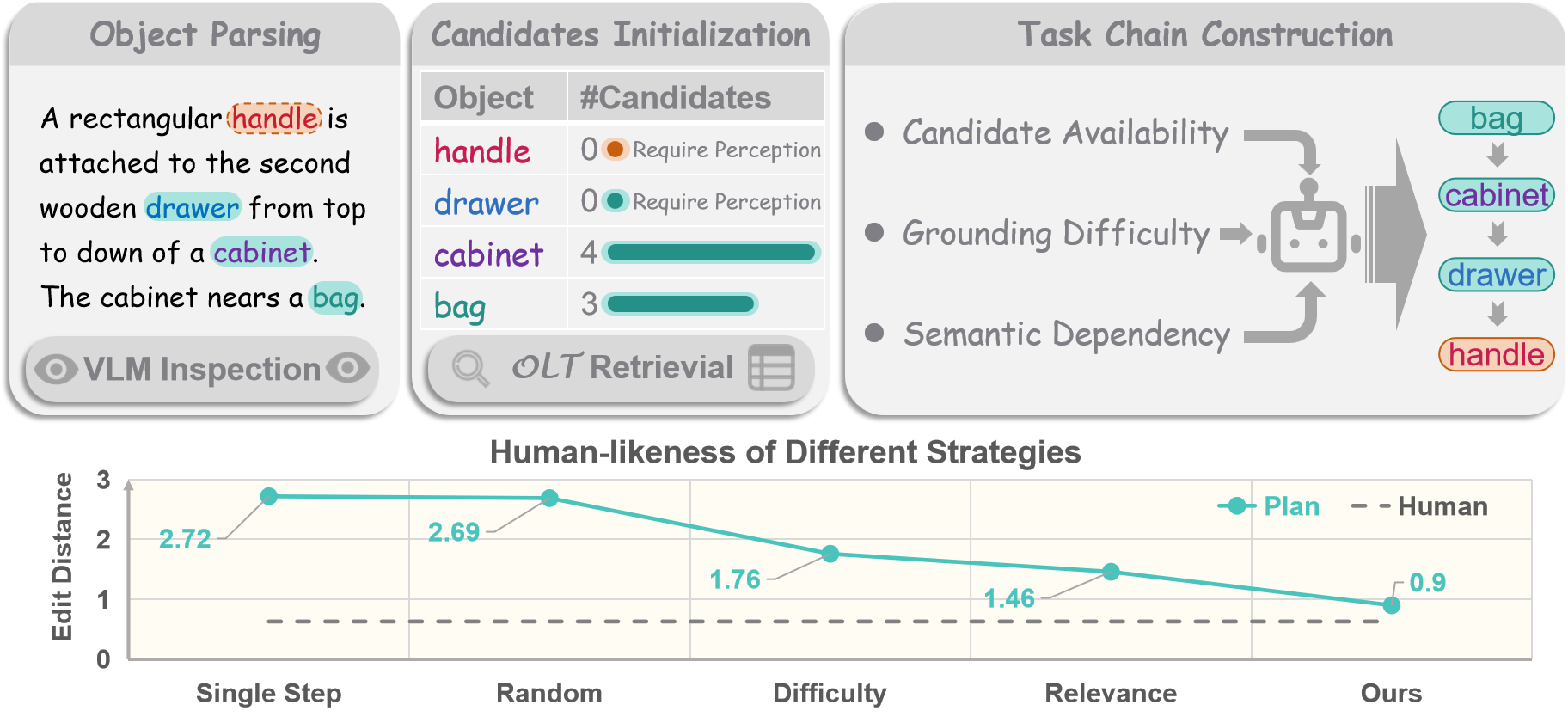}
    \caption{Task-Chain Planning.}
    \label{fig:task_chain_construction_process}
  \end{subfigure}
  \hfill
  \begin{subfigure}[t]{0.49\linewidth}
    \centering
    \includegraphics[width=\linewidth]{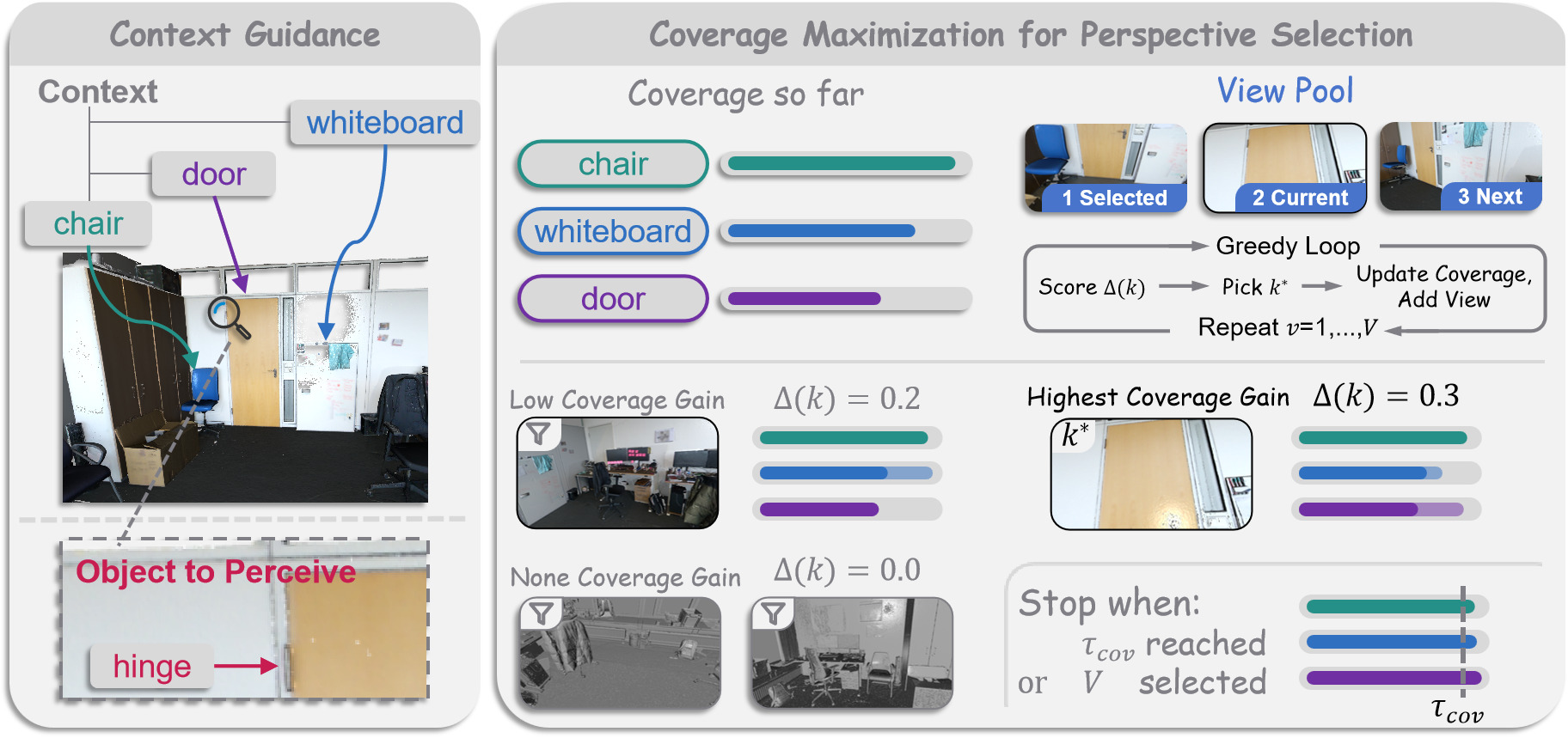}
    \caption{Context-Guided Perspectives Selection.}
    \label{fig:maximize_coverage_display}
  \end{subfigure}

  \myvspace{-8pt}
  \caption{\textbf{Illustrations of key components in OpenGround.} (a): Task-Chain Planning decomposes a query into an ordered sequence of sub-goals. (b): Context-Guided Perspective Selection finds views greedily maximizing context objects coverage.}
  \label{fig:method_illustrations}
  \myvspace{-16pt}
\end{figure}

\subsection{Task-Chain Planning}
\label{sec:task_chain_construction}

When grounding fine-grained objects or in crowded scenarios, queries naturally become complex. In such cases, humans rarely make a one-shot decision; instead, they leverage context and make intermediate confirmations. For example, given query ``find a \textit{handle} of the top \textit{drawer} of the \textit{kitchen cabinet}'', humans would locate the \textit{cabinet} in the \textit{kitchen}, then identify \textit{drawer} and \textit{handle}, instead of selecting the handle from tens of candidates across drawers and rooms. 
Even when human grounding appears parallel, it can be viewed as sequential atomic steps, necessary for deterministic program.
This motivates grounding strategy: decomposing a query into a sequence of sub-goals, where grounded objects narrow the search space for next steps.
However, obtaining a useful strategy is non-trivial. Simple strategies, \eg, ground target directly once an anchor confirmed, using a random order, sorting by difficulty (proxy by the candidate set size), or sorting by semantic relevance, misalign with the dependency structure in complex queries (\eg, attempting a missing target too early or ignoring spatial priors).
We validate this via a user study (settings in \textbf{Appendix}) and find that they produce less human-like (larger edit distance) plans, as illustrated in \cref{fig:task_chain_construction_process}.
Thus, we introduce a novel task chain strategy to obtain human-like plans.

\paragraph{Object Parsing and Candidates Initialization.}
We use a VLM to extract object mentioned in $\query$, including one target label $L^{tgt}$ and a set of relevant context labels $\{L^{ctx}_j\}_{j=1}^{n}$.
For each label in $\mathcal{L}=\{L^{tgt}\}\cup\{L^{ctx}_j\}_{j=1}^{n}$, we initialize its candidate set from the predefined offline $\olt$ using the same candidate retrieval described in \cref{sec:preliminary}. The candidate set size $|\objidset_j|$ serves as a proxy for grounding difficulty, following the difficulty partitioning in Nr3D \cite{achlioptas2020referit3d}.

\paragraph{Task-Chain Construction.}
The core challenge lies in determining an optimal grounding order that respects the implicit context-dependency structure within compositional query. We identify three key factors that influence the ordering:
\begin{enumerate}
    \item \textbf{Candidate Availability}: Objects absent ($|\objidset_j|=0$) require context-guided perception and must be grounded \textit{after} their context-dependency objects.
    \item \textbf{Grounding Difficulty}: Objects with larger candidate sets have broader search spaces and thus higher grounding difficulty. Grounding them later allows previously grounded context objects to narrow the search space.
    \item \textbf{Semantic Dependency}: Objects linked by spatial or possessive relations (\eg, ``handle \textit{of} the drawer'') should be grounded in dependency order.
\end{enumerate}
To capture these factors, we design a structured prompt that inputs: (1) the original query $\query$; (2) the extracted labels $\mathcal{L}$ with their candidate counts; (3) special markers for objects without any candidates, which require Context-Guided Perception. The VLM is instructed to reason over semantic dependencies based on linguistic cues and commonsense, then output an ordered index sequence:
\begin{equation}
     \taskchain = \fvlm\parA{\query, \mathcal{L}, \parB{\|\objidset_j\|\mid j\in [1,n+1]}}.
\end{equation}
Notably, we adopt a soft constraint to encourage grounding the target object last through a prompt, but permits early target grounding when the target possesses strong unique identifiers (\eg, much less candidates, or salient visual features). This flexibility allows adaptively prioritizing the target when necessary.

\paragraph{Validation via User Study.}
To assess alignment with human planning, we compare with the baselines above. Our method achieves an edit distance of $0.90\pm0.49$ to human annotations, outperforming alternative strategiesand approaches the inter-participant agreement level of $0.63\pm0.29$, indicating human-like strategies. Details about the user study can be found in the $\textbf{Appendix}$.

\subsection{Context-Guided Perception}
\label{sec:ace}

Since task-chain planned and previously grounded context objects provide spatial priors, we introduce Context-Guided Perception (CGP) to leverage the context to filter irrelevant views and discover missing objects to extend $\olt$ online.
As illustrated in \cref{fig:overview}, CGP selects informative views under context guidance, applies 2D foundation models to perceive missing objects, lifts them to 3D, and inserts discovered objects into $\olt$.
This converts the offline OLT into an online OLT while preserving the same downstream interface for single-step grounding. 

\paragraph{Context-Guided Perspective Selection.}
Selecting observation views is crucial for efficient and accurate perception. Existing 3DVG methods rely on heuristics that either underuse contextual cues or waste views: SeeGround \cite{li2025seeground} uses a single geometry-derived view that often misses task-relevant regions or yields distant views with insufficient detail; SeqVLM \cite{lin2025seqvlm} favors zoomed-in views where each candidate dominates the image, producing redundant views and losing context; Struct2D \cite{zhu2025struct2d} uses bird's-eye views that suffer from occlusion and miss side details. These approaches share a common limitation: they lack awareness of complex task structure and cannot leverage the progressively grounded context from task-chain to focus on where the missing object is most likely to appear.

Our key insight is that grounded context objects $[O_1,\cdots,O_{t-1}]$ serve as spatial anchors for task-relevant regions implied by the query's compositional relations. 
We thus select views that maximize coverage of these context objects, which exposes nearby regions where the missing object is likely to appear.

Specifically, for the current task $T_t$ with missing candidates, we select views $\rgbset^\star\subseteq\rgbset$ around $[O_1,\cdots,O_{t-1}]$ as context guidance.
We prefer views that increase coverage of these context objects, as illustrated in \cref{fig:maximize_coverage_display}.
We adopt a greedy strategy to select at most $V$ perspectives (global optimization is NP-hard), inspired by Video-LLM \cite{zheng2025video} which uses greedy strategy to sample views to obtain maximum coverage of full scene.
For each object $O_i$, initialize $\mathcal{R}^0_{O_i}=\emptyset$ and iteratively add views for $v=1,\cdots,V$.
For each view $k$, coverage gain is:
\begin{equation}
    \Delta(k) = \sum_{O_i}\frac{|\pcd_{O_i}\cap\pcd_k\backslash\mathcal{R}^{v-1}_{O_i}|}{|\pcd_{O_i}|},
\end{equation}
where $\pcd_{O_i}$ is the point set of $O_i$, $\pcd_k$ denotes the 3D points visible from view $k$, and $\mathcal{R}^{v-1}_{O_i}$ denotes the observed region after $v-1$ selected perspectives.
At step $v$, we select $k^\star=\arg\max_k \Delta(k)$, add $\rgb_{k^\star}$ into $\rgbset^\star$, and update:
\begin{equation}
    \mathcal{R}^{v}_{O_i}=\mathcal{R}^{v-1}_{O_i}\cup\parA{\pcd_{O_i}\cap\pcd_{k^\star}}, \forall O_i.
\end{equation}
The summation encourages balanced coverage across all context objects. The process stops when coverage reaches $\tau_{cov}$ or $V$ views are selected. With context-guided views $\rgbset^\star$ ready, we proceed to detect missing objects via following.

\paragraph{2D Segmentation and Lifting.}
Given $\rgbset^\star$, we detect missing objects in each image with an open-world 2D segmentor and lift the masks to 3D to extend the $\olt$.
For each view $k$, we obtain $\maskset_k=\fseg(\rgb_k, L_{T_t})$, and lift them to 3D using the corresponding point cloud and camera parameters.
We aggregate lifted masks across views and iteratively merge those with high overlap (3D IoU $\ge \tau_{iou}$) to obtain distinct 3D instances $\{\pcd_i\}_{i=1}^{M}$, following existing work \cite{nguyen2024open3dis}.
Finally, we covert each object to a 3D bounding box and insert to $\olt$:
\begin{equation}
    \olt \leftarrow \olt \cup \Big\{ \big(\mathbf{ID}(i),\, L_{T_t},\, \bbox(\pcd_i)\big)\ \big|\ i\in[1,M]\Big\},
\end{equation}
where $\mathbf{ID}(i)$ denotes the next unique id and $\bbox(\pcd_i)$ is the bounding box.

In summary, Task-Chain Planning orchestrates the pipeline by processing $\taskchain=[T_1,\cdots,T_{n+1}]$ sequentially: for each $T_t$, if candidates exist in $\olt$, directly apply single-step grounding; otherwise, CGP uses previously grounded objects to discover missing objects, extends $\olt$ online, and then applies single-step grounding. Each resolved sub-goal progressively narrows the search space, enabling OpenGround to handle complex queries and novel objects in open-world. 


\section{Experiments}
\label{sec:exp}

\subsection{Experimental Settings}

\paragraph{Datasets.} We evaluate OpenGround on three datasets to assess both open-world and OLT-dependent grounding performance. For open-world evaluation, we conduct experiments on all 7,724 queries of OpenTarget (details in \cref{sec:dataset}), which contains part-level objects beyond usual categories to simulate the open-world scenarios, to test the ability of grounding open-world objects. For OLT-dependent evaluation, we use two popular benchmarks: ScanRefer \cite{chen2020scanrefer} and Nr3D \cite{achlioptas2020referit3d}. ScanRefer provides 9,508 queries in validation set, on ScanNet \cite{dai2017scannet} scenes. Nr3D, part of the ReferIt3D benchmark \cite{achlioptas2020referit3d}, includes 7,805 queries in validation set and offers ground-truth 3D bounding boxes for objects.

\paragraph{Implementation Details.} Our experiments utilize the GLM-4.5V \cite{vteam2025glm45v} as the VLM for OpenTarget evaluation, CLIP-ViT-Base-Patch16 text encoder \cite{radford2021clip} for text similarity matching, GroundedSAM \cite{ren2024grounded} as $\fseg$ and set $V$=3, $\tau_{cov}$=0.95, $\tau_{iou}$=0.5 where $\tau_{cov}$ and $\tau_{iou}$ follow existing works \cite{zheng2025video,nguyen2024open3dis}. For the $\olt$ on ScanRefer and OpenTarget benchmarks, we follow the object detection procedure outlined in ZSVG3D \cite{yuan2024visual} for fair comparison, which utilizes Mask3D \cite{schult2022mask3d}.

\subsection{Comparative Study}

We conduct experiments across benchmarks, including Nr3D \cite{achlioptas2020referit3d}, ScanRefer\cite{chen2020scanrefer}, and OpenTarget. We present detailed performance on Nr3D and ScanRefer, with category-wise breakdowns in \cref{tab:scanrefer_comparison,tab:nr3d_comparison}, and report OpenTarget comparisons in \cref{tab:opentarget_comparison}. Due to time constraints, we only evaluate the performance of recently proposed top-performing and open-source zero-shot methods on OpenTarget.

\begin{figure}
    \centering
    \includegraphics[width=0.9\linewidth]{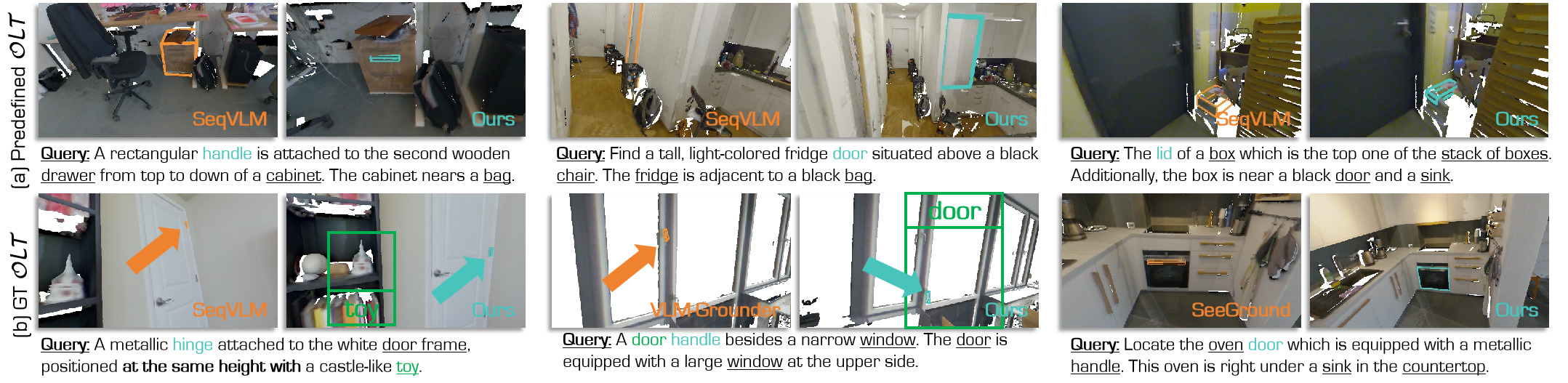}
    \caption{Qualitative Comparisons on \textbf{OpenTarget}. (a) compares our method with previous open-source SoTA method, SeqVLM \cite{lin2025seqvlm} with predefined $\olt$. They fail to ground objects out-of-$\olt$. (b) compares our method with previous methods equipped with ground-truth $\olt$. these baselines either overlook key objects (\eg, toy, door) or are distracted by numerous relevant objects, leading to failures.}
    \label{fig:qualitative_comp}
\end{figure}

\begin{table}
    \centering
    \caption{Performance on \textbf{OpenTarget}. $^*$ denotes results on randomly selected 300 samples due to low efficiency. Methods fail due to missing objects in the Mask3D $\olt$. \textbf{Easy} contains queries with hierarchy length $\leq 2$ (5,737 samples), while \textbf{Hard} contains queries with hierarchy length $> 2$ (1,987 samples).}
    \scalebox{0.75}{
        \begin{tabular}{l|c|cc|cc|cc}
            \toprule
            \multirow{2}{*}{Method} & \multirow{2}{*}{$\olt$}  & \multicolumn{2}{c|}{Easy} & \multicolumn{2}{c|}{Hard} & \multicolumn{2}{c}{Overall} \\
            \cmidrule{3-4} \cmidrule{5-6} \cmidrule{7-8}
            & & Acc@0.25 & Acc@0.50 & Acc@0.25 & Acc@0.50 & Acc@0.25 & Acc@0.50 \\
        \midrule
        \midrule
        SeeGround \cite{li2025seeground} & Mask3D \cite{schult2022mask3d} & 13.3 & 11.2 & 1.2 & 1.1 & 10.2 & 8.6 \\
        VLM-Grounder$^*$ \cite{xu2024vlmgrounder} & - & 14.5 & 12.1 & 6.1 & 2.3 & 12.3 & 9.6 \\
        VLM-Grounder$^*$ \cite{xu2024vlmgrounder} & Mask3D \cite{schult2022mask3d} & 16.4 & 12.0 & 10.2 & 4.8 & 14.8 & 10.1 \\
        SeqVLM \cite{lin2025seqvlm} & Mask3D \cite{schult2022mask3d} & 13.6 & 11.2 & 1.4 & 1.1 & 10.5 & 8.6 \\
        GPT4Scene \cite{qi2025gpt4scene} & Mask3D \cite{schult2022mask3d} & 10.1 & 7.7 & 0.9 & 0.7 & 7.7 & 5.9 \\
        ZSVG3D \cite{yuan2024visual} & Mask3D \cite{schult2022mask3d} & 8.9 & 6.9 & 0.8 & 0.7 & 6.8 & 5.3 \\
        \textbf{Ours} & Mask3D \cite{schult2022mask3d} & \textbf{51.8} & \textbf{38.9} & \textbf{30.2} & \textbf{20.6} &\textbf{46.2} & \textbf{34.2} \\
        \midrule
        SeeGround \cite{li2025seeground} & GT & 20.2 & 19.8 & 11.3 & 10.4 & 17.9 & 17.4 \\
        VLM-Grounder$^*$ \cite{xu2024vlmgrounder} & GT & 31.4 & 21.3 & 20.6 & 17.8 & {28.6} & {20.4} \\
        SeqVLM \cite{lin2025seqvlm} & GT & 21.5 & 21.3 & 13.4 & 13.2 & 19.4 & 19.2 \\
        GPT4Scene \cite{qi2025gpt4scene} & GT & 13.6 & 13.4 & 7.9 & 7.3 & 12.1 & 11.8 \\
        ZSVG3D \cite{yuan2024visual} & GT & 12.2 & 12.0 & 7.6 & 6.8 & 11.0 & 10.7 \\
        \textbf{Ours} & GT & \textbf{57.9} & \textbf{57.4} & \textbf{45.7} & \textbf{45.3} & \textbf{54.8} & \textbf{54.3} \\
        \bottomrule
    \end{tabular}
    }
    \label{tab:opentarget_comparison}
\end{table}
\paragraph{OpenTarget.} As illustrated in \cref{tab:opentarget_comparison}, on the open-world benchmark, existing methods are constrained by predefined offline OLT which inevitably lacks the target objects, resulting in severely degraded performance (visualized in \cref{fig:qualitative_comp}(a)). To fairly assess their upper bound, according to the conclusion in relevant works that better $\olt$ results better performance, we further evaluate them with ground-truth (GT) $\olt$. However, they are still confounded by the larger candidate pool, resulting in erroneous predictions visualized in \cref{fig:qualitative_comp}(b) and reported in \cref{tab:opentarget_comparison}. In contrast, our method far surpasses baselines although they are given GT $\olt$. It shows that OpenGround effectively extends the $\olt$ with novel objects and resolves complex queries, enabling open-world grounding.

\begin{table}[t]
    \centering
    \caption{Quantitative Comparisons on \textbf{ScanRefer} \cite{chen2020scanrefer}. Results are reported for “Unique” (scenes with a single target object) and “Multiple” (scenes with distractors of the same class) subsets. $^*$ denotes results on selected 250 samples.} 
    \scalebox{0.75}{
        \begin{tabular}{l|c|c|cc|cc|cc}
            \toprule
            \multirow{2}{*}{Method} & \multirow{2}{*}{Supervision} & \multirow{2}{*}{VLM} & \multicolumn{2}{c|}{Unique} & \multicolumn{2}{c|}{Multiple} & \multicolumn{2}{c}{Overall} \\
            \cmidrule{4-5} \cmidrule{6-7} \cmidrule{8-9}
            & & & Acc@0.25 & Acc@0.50 & Acc@0.25 & Acc@0.50 & Acc@0.25 & Acc@0.50 \\
            \midrule \midrule
            ViewSRD \cite{huang2025viewsrd} & Supervised & - & 82.1 & 68.2 & 37.4 & 29.0 & 45.4 & 36.0 \\
            3D-R1 \cite{huang20253d} & Supervised & - & - & - & - & - & \textbf{65.8} & \textbf{59.2} \\
            GPT4Scene \cite{qi2025gpt4scene} & Supervised & - & 90.3 & 83.7 & 56.4 & 50.9 & \uline{62.6} & \uline{57.0} \\
            TSP3D \cite{guo2025text} & Supervised & - & - & - & - & - & 56.4 & 46.7 \\
            \midrule \midrule
            SeeGround \cite{li2025seeground} & Zero-Shot & Qwen2-VL-72b \cite{Qwen2-VL} & 75.7 & 68.9 & 34.0 & 30.0 & 44.1 & 39.4 \\
            \textbf{Ours} & Zero-Shot &  Qwen2-VL-72b \cite{Qwen2-VL} & \textbf{76.6} & \textbf{70.3} & \textbf{48.1} & \textbf{40.4} & \textbf{53.7} & \textbf{46.3} \\
            \midrule
            SeqVLM \cite{lin2025seqvlm} & Zero-Shot & Doubao-1.5-pro \cite{doubao} & 77.3 & 72.7 & 47.8 & 41.3 & 55.6 & 49.6 \\
            \textbf{Ours} & Zero-Shot &  Doubao-1.5-pro \cite{doubao} & \textbf{78.3} & \textbf{75.1} & \textbf{59.2} & \textbf{49.4} & \textbf{63.0} & \textbf{54.4} \\
            \midrule
            VLM-Grounder$^*$ \cite{xu2024vlmgrounder} & Zero-Shot & GPT-4o \cite{openai2024gpt4technicalreport} & 51.6 & 32.8 & \textbf{66.0} & 29.8 & 48.3 & 33.5 \\
            SPAZER \cite{jin2025spazer} & Zero-Shot & GPT-4o \cite{openai2024gpt4technicalreport} & 80.9 & 72.3 & 51.7 & 43.4 & 57.2 & 48.8 \\
            \textbf{Ours} & Zero-Shot &  GPT-4o \cite{openai2024gpt4technicalreport} & \textbf{83.5} & \textbf{75.2} & \uline{59.9} & \textbf{51.2} & \textbf{64.5} & \textbf{55.9} \\
            \midrule
            ZSVG3D \cite{yuan2024visual} & Zero-Shot & GPT-4 turbo \cite{openai2024gpt4technicalreport} & 63.8 & 58.4 & 27.7 & 24.6 & 36.4 & 32.7 \\
            \textbf{Ours} & Zero-Shot &  GPT-4 turbo \cite{openai2024gpt4technicalreport} & \textbf{78.0} & \textbf{75.8} & \textbf{58.5} & \textbf{48.2} & \textbf{62.3} & \textbf{53.6} \\
            \bottomrule
        \end{tabular}
    }
    \label{tab:scanrefer_comparison}
\end{table}
\begin{table}[t]
    \centering
    \caption{Detailed Performance on \textbf{Nr3D} \cite{achlioptas2020referit3d}. Queries are categorized as “Easy” (with one distractor) or “Hard” (with distractors), and as “Dep.” (View-Dependent) or “Indep.” (View-Independent) based on viewpoint requirements for grounding. ``\textbf{Ours}+SPAZER \cite{jin2025spazer}'' seamlessly integrates its advanced module into our framework.}
    \scalebox{0.75}{
    \begin{tabular}{l|c|c|cc|cc|c}
        \toprule
        Method & Supervision & VLM & Easy & Hard & Dep. & Indep. & Overall \\
        \midrule
        \midrule
        3D-R1 \cite{huang20253d} & Supervised & - & - & - & - & - & \uline{68.8} \\
        TSP3D \cite{guo2025text} & Supervised & - & - & - & - & - & 48.7 \\
        ViewSRD \cite{huang2025viewsrd} & Supervised & - & 75.3 & 64.8 & 68.6 & 70.6 & \textbf{69.9} \\
        \midrule
        \midrule
        SeeGround \cite{li2025seeground} & Zero-Shot & Qwen2-VL-72b \cite{Qwen2-VL} & 54.5 & 38.3 & 42.3 & 48.2 & 46.1 \\
        \textbf{Ours} & Zero-Shot & Qwen2-VL-72b \cite{Qwen2-VL} & \textbf{54.9} & \textbf{44.6} & \textbf{46.4} & \textbf{52.9} & \textbf{50.6} \\
        \midrule
        SeqVLM \cite{lin2025seqvlm} & Zero-Shot & Doubao-1.5-pro \cite{doubao}  & 58.1 & 47.4 & 51.0 & 54.5 & 53.2 \\
        \textbf{Ours} & Zero-Shot & Doubao-1.5-pro \cite{doubao} & \textbf{63.8} & \textbf{57.4} & \textbf{58.9} & \textbf{61.4} & \textbf{60.5} \\
        \midrule
        VLM-Grounder \cite{xu2024vlmgrounder} & Zero-Shot & GPT-4o \cite{openai2024gpt4technicalreport} & 55.2 & 39.5 & 45.8 & 49.4 & 48.0 \\
        SPAZER \cite{jin2025spazer} & Zero-Shot & GPT-4o \cite{openai2024gpt4technicalreport} & \textbf{68.0} & \uline{58.8} & \textbf{59.9} & \textbf{66.2} & \textbf{63.8} \\
        \textbf{Ours}  & Zero-Shot & GPT-4o \cite{openai2024gpt4technicalreport} &\uline{64.3} & \textbf{59.3} & \uline{59.2} & \uline{63.1} & \uline{61.7} \\
        \textbf{Ours}+SPAZER \cite{jin2025spazer} & Zero-Shot & GPT-4o \cite{openai2024gpt4technicalreport} & \textbf{70.1} & \textbf{59.8} & \textbf{60.4} & \textbf{67.2} & \textbf{64.8}\\
        \midrule
        ZSVG3D \cite{yuan2024visual} & Zero-Shot & GPT-4 turbo \cite{openai2024gpt4technicalreport} & 46.5 & 31.7 & 36.8 & 40.0 & 39.0 \\
        \textbf{Ours}  & Zero-Shot & GPT-4 turbo \cite{openai2024gpt4technicalreport} & \textbf{59.3} & \textbf{55.1} & \textbf{55.1} & \textbf{58.2} & \textbf{57.1} \\
        \bottomrule
    \end{tabular}
    }
    \label{tab:nr3d_comparison}
\end{table}
\paragraph{Nr3D \& ScanRefer.}
OpenGround achieves competitive performance across both benchmarks. On Nr3D \cite{achlioptas2020referit3d}, we achieve competitive accuracy to the zero-shot SoTA SPAZER \cite{jin2025spazer}. On ScanRefer \cite{chen2020scanrefer}, we outperform all zero-shot baselines and narrow the gap to supervised methods. These improvements are mainly incremental rather than dramatic, because the evaluation settings are simpler: queries are shorter, and scenes contain fewer candidates, which reduces the need for task-chain planning, shown in \cref{sec:dataset}. Moreover, targets in Nr3D and ScanRefer are largely covered by the predefined OLT, so our context-guided perception is rarely used and thus contributes little under these close-world conditions.

\subsection{Ablation Study}

To validate each module in OpenGround, we conduct ablation studies on the OpenTarget. All experiments use the same settings as the main results.

\paragraph{Effect of Initial $\olt$.} As shown in \cref{tab:ablation_comparison}, OpenGround remains effective even without an initial $\olt$, outperforming baselines with GT $\olt$. Nevertheless, incorporating an initial $\olt$ offers priors that further improves performance.

\begin{table}[t]
  \centering
  \caption{Ablation study of different components on OpenTarget Acc@0.50. \textbf{Initial }$\olt$: whether using initial $\olt$ from Mask3D \cite{schult2022mask3d}; \textbf{Grounding Strategy}: strategy for constructing grounding sequence; \textbf{VLM}: the vision-language model used. \textbf{Right:} swap only the Single-Step Grounding backbone within OpenGround (others fixed).}
  \label{tab:ablation_comparison}

  \begin{minipage}[t]{0.48\linewidth}
    \vspace{0pt}
    \centering
    \scalebox{0.75}{
      \begin{tabular}{l|ccc|c}
        \toprule
        \# & \makecell{Initial\\$\olt$} & \makecell{Grounding\\Strategy} & VLM & Acc \\
        \midrule
        (1) & \text{Yes} & \text{Task Chain} & \text{GLM-4.5V} \cite{vteam2025glm45v} & 34.2 \\
        \midrule
        (2) & \textit{No} & \text{Task Chain} & \text{GLM-4.5V} \cite{vteam2025glm45v} & 27.1 \\
        \midrule
        (3) & \text{Yes} & \textit{Jump} & \text{GLM-4.5V} \cite{vteam2025glm45v} & 29.8\\
        (4) & \text{Yes} & \textit{Relevance} & \text{GLM-4.5V} \cite{vteam2025glm45v} & 32.6 \\
        (5) & \text{Yes} & \textit{Difficulty} & \text{GLM-4.5V} \cite{vteam2025glm45v} & 31.5 \\
        (6) & \text{Yes} & \textit{Random} & \text{GLM-4.5V} \cite{vteam2025glm45v} & 29.2 \\
        \midrule
        (7) & \text{Yes} & \text{Task Chain} & \textit{Qwen3-VL-32B} \cite{qwen3technicalreport} & 30.4 \\
        (8) & \text{Yes} & \text{Task Chain} & \textit{Qwen3-VL-235B} \cite{qwen3technicalreport} & 32.8 \\
        (9) & \text{Yes} & \text{Task Chain} & \textit{Step3} \cite{step3system} & 33.4 \\
        \bottomrule
      \end{tabular}
    }
  \end{minipage}
  \begin{minipage}[t]{0.50\linewidth}
    \vspace{0pt}
    \centering
    \includegraphics[width=\textwidth]{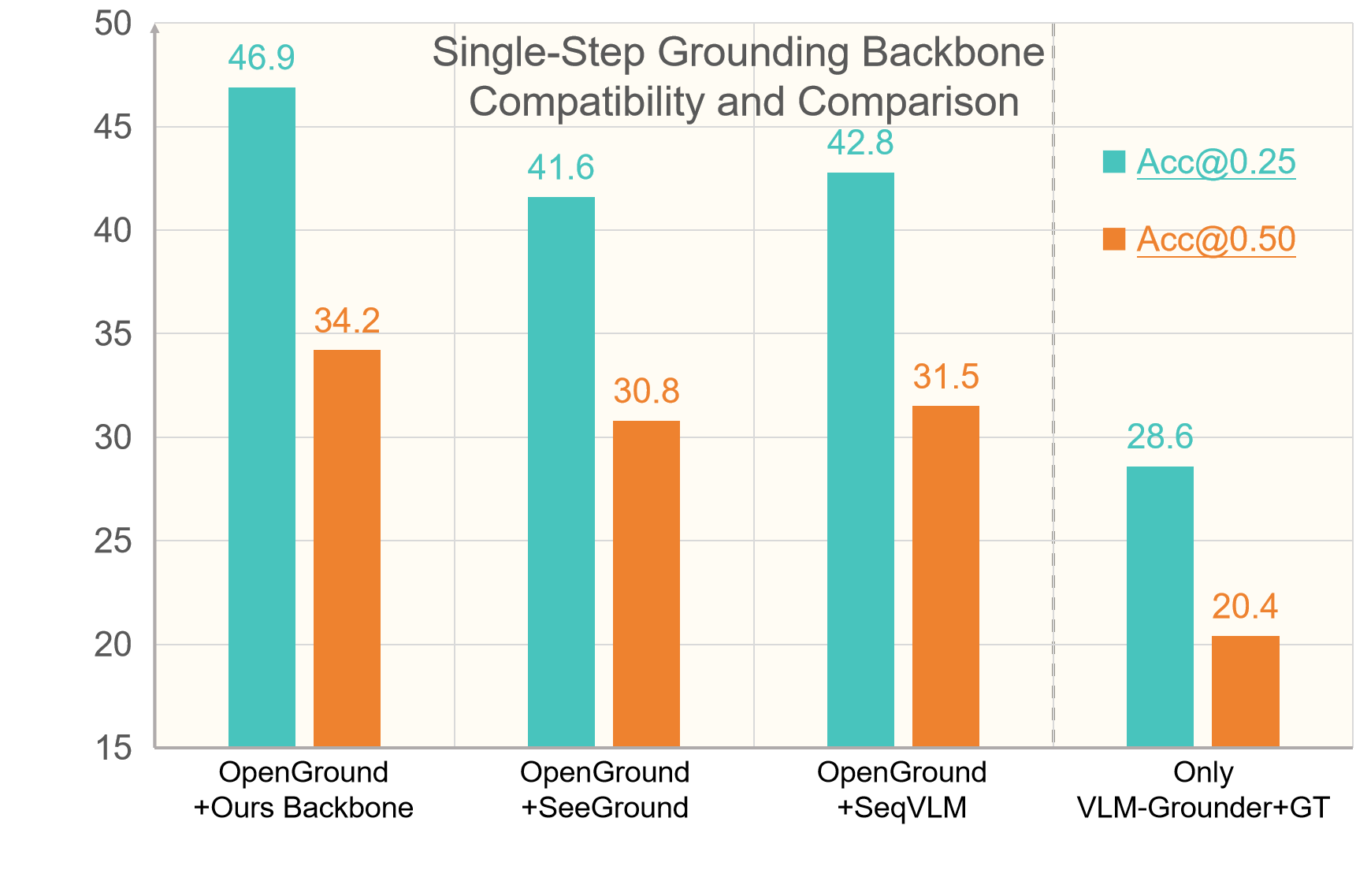}
  \end{minipage}
\end{table}

\begin{figure}[h]
    \centering
    \includegraphics[width=\linewidth]{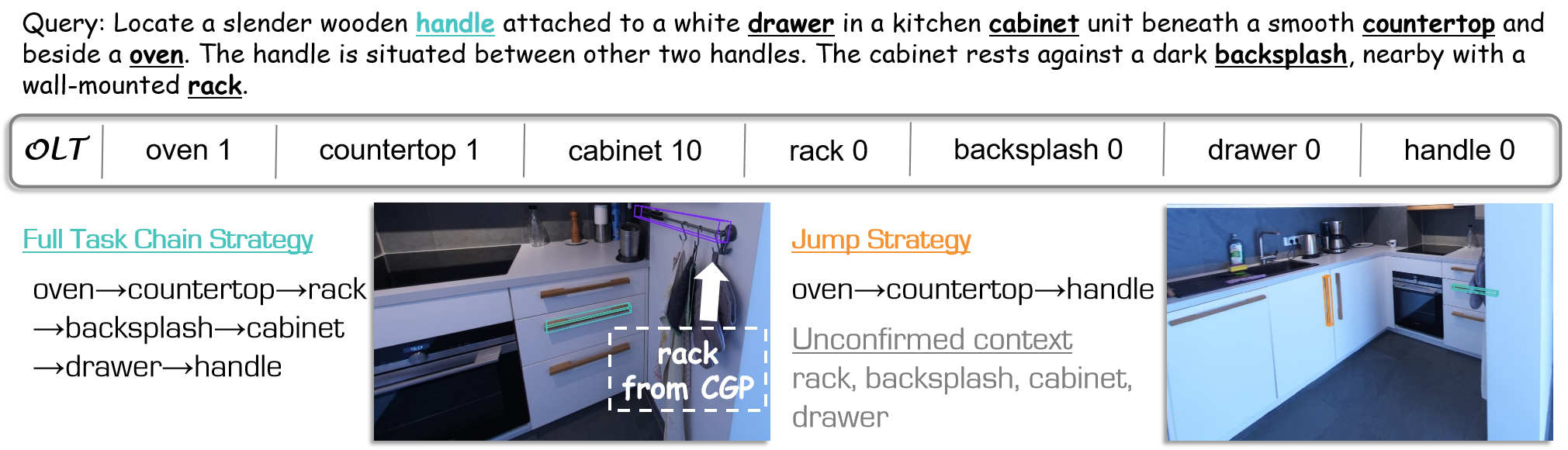}
    \caption{\textbf{Grounding Strategy Results Comparison.} \textbf{Jump} strategy incorrectly identifies the object (orange) because it skips the key object, rack. \textbf{Task Chain} strategy correctly identifies the object (teal), considering the rack (purple).}
    \label{fig:task_chain_result_comp}
\end{figure}

\paragraph{Task Chain Construction Strategy.}
We compare several grounding plan construction strategies (\cref{tab:ablation_comparison}, rows 1, 3--6): \textbf{Random} uses a random order; \textbf{Jump} grounds target directly once an object confirmed; \textbf{Difficulty} prioritizes easier objects by candidate set size; \textbf{Semantic} orders by relevance to the target; 
\textbf{Task Chain} is our strategy and achieves the best performance.
\cref{fig:task_chain_result_comp} further compares with \textbf{Jump}, where ours yields more coherent intermediate grounding and more accurate grounding via progressive planning. 
Together with  \cref{fig:task_chain_construction_process}, these results demonstrate that human-like planning improves open-world grounding.

\paragraph{Influence of VLM.} To verify VLM dependence and adaptability, we replace the default GLM-4.5V \cite{vteam2025glm45v} with VLMs of varying scales. As shown in \cref{tab:scanrefer_comparison,tab:nr3d_comparison,tab:ablation_comparison}, OpenGround’s advantage stems from our contributions rather than large-scale VLM capabilities. Notably, smaller VLMs induces minimal performance degradation, even that the Qwen3-VL-32B \cite{qwen3technicalreport} only drops 3.8\%, still outperforms baselines that leverage GT $\olt$ (\eg, 10\% higher than VLM-Grounder).

\begin{figure}
  \centering
  \begin{subfigure}[t]{0.32\linewidth}
    \centering
    \includegraphics[width=\linewidth]{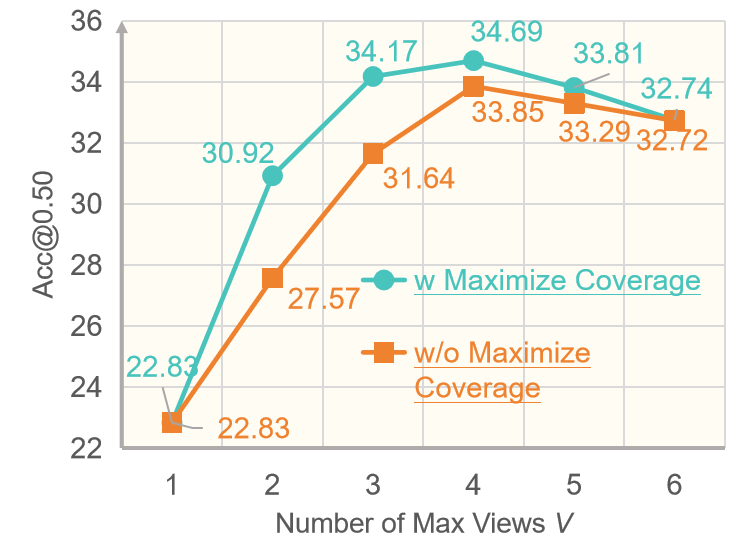}
    \caption{Perspective selection vs. $V$.}
    \label{fig:perspectives_selection_comp}
  \end{subfigure}
  \hfill
  \begin{subfigure}[t]{0.32\linewidth}
    \centering
    \includegraphics[width=\linewidth]{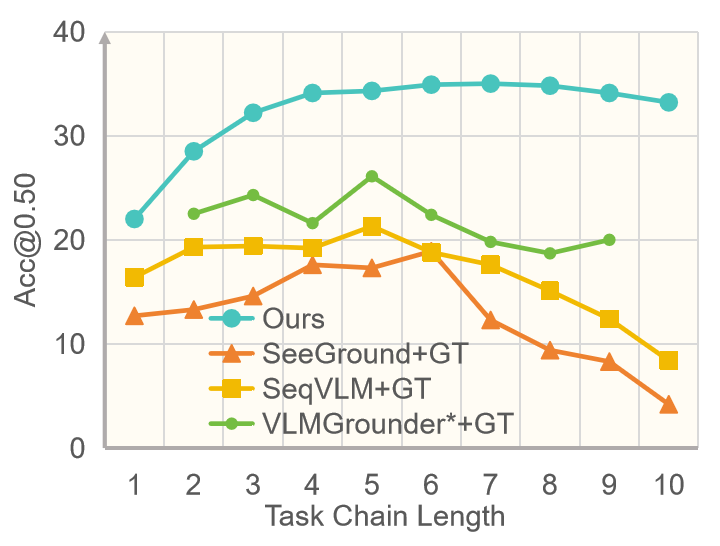}
    \caption{Accuracy vs. chain length.}
    \label{fig:task_chain_quantitative_acc}
  \end{subfigure}
  \hfill
  \begin{subfigure}[t]{0.32\linewidth}
    \centering
    \includegraphics[width=\linewidth]{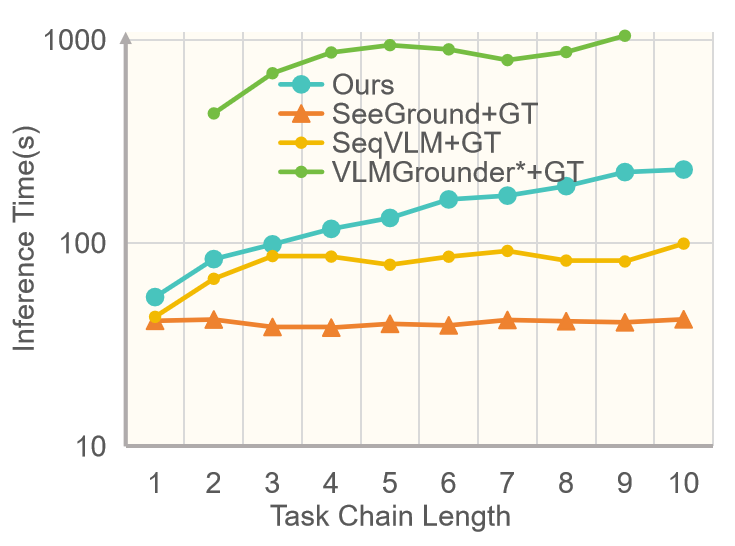}
    \caption{Time vs. chain length.}
    \label{fig:task_chain_quantitative_time}
  \end{subfigure}
  \caption{\textbf{Ablation studies.} (a): impact of perspective selection strategies with varying max views $V$. (b) and (c): performance and efficiency across query complexity which is measured by task chain length our method planned. ``VLMGrounder${}^*$+GT'' is evaluated only on a subset of 300 samples due to its extremely high computational cost, and lacks values at $L=1$ and $L=10$ since these lengths do not appear in the sampled subset.}
  \label{fig:ablation_combined}
\end{figure}

\paragraph{Single-Step Grounding Backbones.} We evaluate the compatibility of our framework with existing methods. As shown in \cref{tab:ablation_comparison}, the single-step grounding module can be seamlessly replaced with prior methods such as SeeGround \cite{li2025seeground} and SeqVLM \cite{lin2025seqvlm}. Results show that CGP enables their open-world grounding ability and Task-Chain Planning boosts performance, confirming that OpenGround is a general and flexible framework rather than a standalone method.

\paragraph{Perspectives Selection Strategy.} 
As illustrated in \cref{fig:perspectives_selection_comp}, we further study the impact of the perspectives selection strategy:
\begin{itemize}
    \item \textbf{Number of views }$V$\textbf{: } When $V=1$, both strategies start at the same low baseline. As $V$ increases to 4, performance of both strategies rises significantly and peaks at 34.7 and 33.8. Exceeding $V=4$ leads to performance drops, indicating that excessive views introduce redundancy and confusion. Though $V=4$ is most accurate, it only brings a 0.52\% gain but increases input views by 33.3\%. We adopt $V=3$, balancing performance and efficiency.
    \item \textbf{Maximizing Coverage Strategy: } The `w Maximize Coverage' outperforms across all $V$ values. This demonstrates that maximizing coverage in view selection avoids CGP missing critical cues, resulting in higher accuracy.
\end{itemize}

\paragraph{Query Complexity Robustness.} As illustrated in \cref{fig:task_chain_quantitative_acc,fig:task_chain_quantitative_time}, our method is robust when query complexity grows, only costing expected linear time.

\paragraph{Limitation.} Our method is designed for static scenes and assumes spatially proximal relevant objects, which are common in current 3DVG. Additionally, CGP’s performance depends on the segmentation model, which lies beyond the scope of our research which focuses on open-world 3DVG. These aspects are left for future exploration. More details about segmentation model influence and possible additional mechanism and error propagation discussion about task chain can be found in \textbf{Appendix}.
\section{Conclusion}


This paper presents OpenGround, a zero-shot framework for open-world 3D visual grounding. Unlike prior methods that either exhaustively complete the object lookup table ($\olt$) or abandon it entirely, OpenGround introduces two modules: Task-Chain Planning that decomposes complex queries into ordered sub-goals for progressive grounding, and Context-Guided Perception that extends the $\olt$ online to discover novel objects on demand. We also contribute the OpenTarget benchmark with 7,724 object-query pairs for open-world evaluation. Experiments show that OpenGround remains competitive on standard datasets (Nr3D, SceneRefer) and achieves state-of-the-art performance on OpenTarget, demonstrating its flexibility and capability in open-world scenarios.

\section*{Acknowledgments}

This work was supported by NSFC under Grant No. 62376121, Basic Research Program of Jiangsu under Grant No. BK20251999, Gusu Innovation Leading Talent Program under Grant No. ZXL2025319, Jiangsu Provincial Science \& Technology Major Project under Grant No. BG2024042, and funded by Nanjing University-China Mobile Communications Group Co.,Ltd. Joint Institute.

%
%
\bibliographystyle{splncs04}
\bibliography{main}

\clearpage

\section{Dataset Details}

\begin{table*}
    \centering
    \caption{\textbf{Sentence Type Examples.} Attributes: (1-6); parts: (2-6); comparatives: (3), (5); superlatives: (4); inter-class spatial relations: (1-6); intra-class spatial relations: (3), (6); ordinal numbers: (6).}
    \begin{tabularx}{\linewidth}{c|X}
    \toprule
        (1) & It is a small black rectangular panel mounted on a vertical pipe beside a white door. \\
        (2) & It is the lid of a small black rectangular panel mounted on a vertical pipe beside a white door.\\
        (3) & It is a metallic hinge, the upper one, mounted on a white door frame. The door features a metallic handle and a closing mechanism at the top, positioned near a ladder and storage items. \\
        (4) & It is a small, round metallic knob mounted on the rightmost door of a tall, light-wood cabinet with four doors. The cabinet stands against the wall near a blue door, beneath exposed ceiling ductwork. \\
        (5) & A metallic hinge attached to the white door frame, positioned at the same height with a castle-like toy. \\
        (6) & Locate the silver handle on the second drawer from the top of a white four-drawer cabinet. The cabinet sits beside a blue office chair, under a small desk with an open cardboard box, near a floral couch. \\
    \bottomrule
    \end{tabularx}
    \label{tab:sentence_type_examples}
\end{table*}

\begin{figure}
    \centering
    \begin{subfigure}{0.49\linewidth}
        \includegraphics[width=\linewidth]{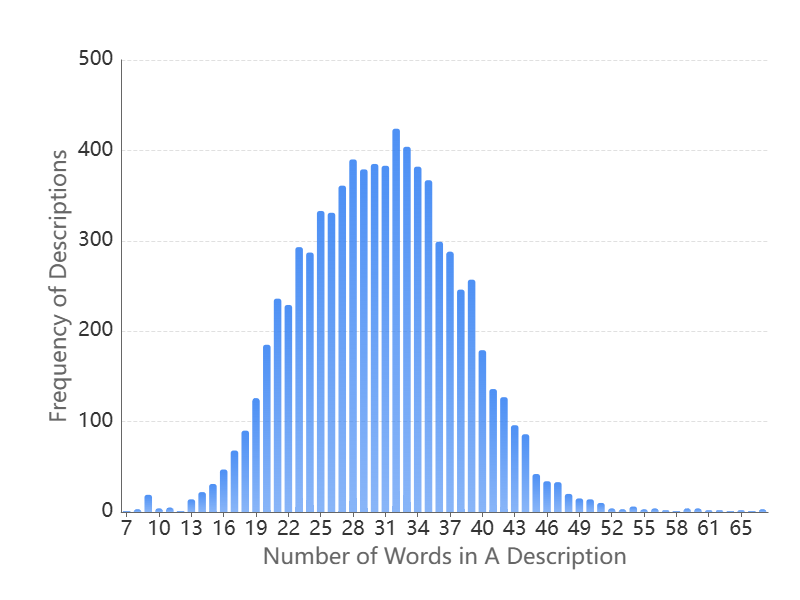}   
        \caption{Description lengths}
        \label{fig:description_lengths}
    \end{subfigure}
    \begin{subfigure}{0.49\linewidth}
        \includegraphics[width=\linewidth]{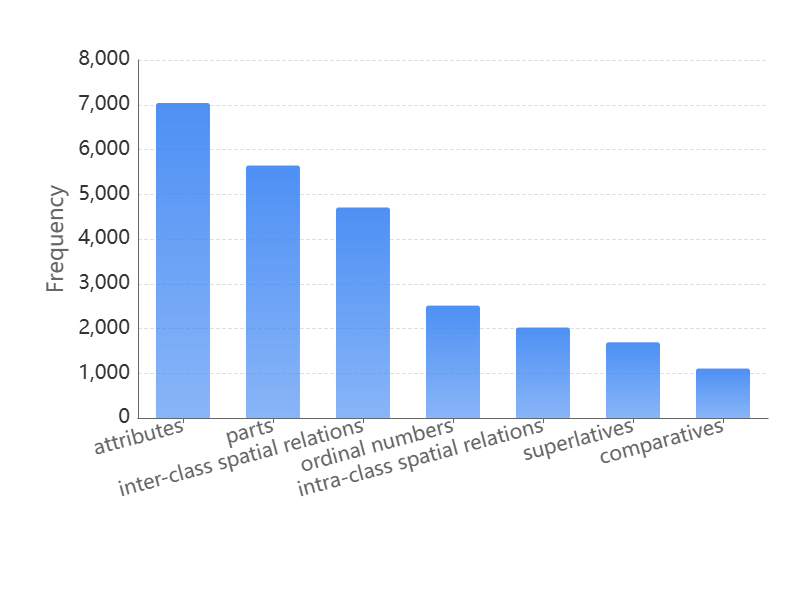}   
        \caption{Sentence Type Distribution}
        \label{fig:sentence_distribution}
    \end{subfigure}
      \begin{subfigure}{0.49\linewidth}
        \includegraphics[width=\linewidth]{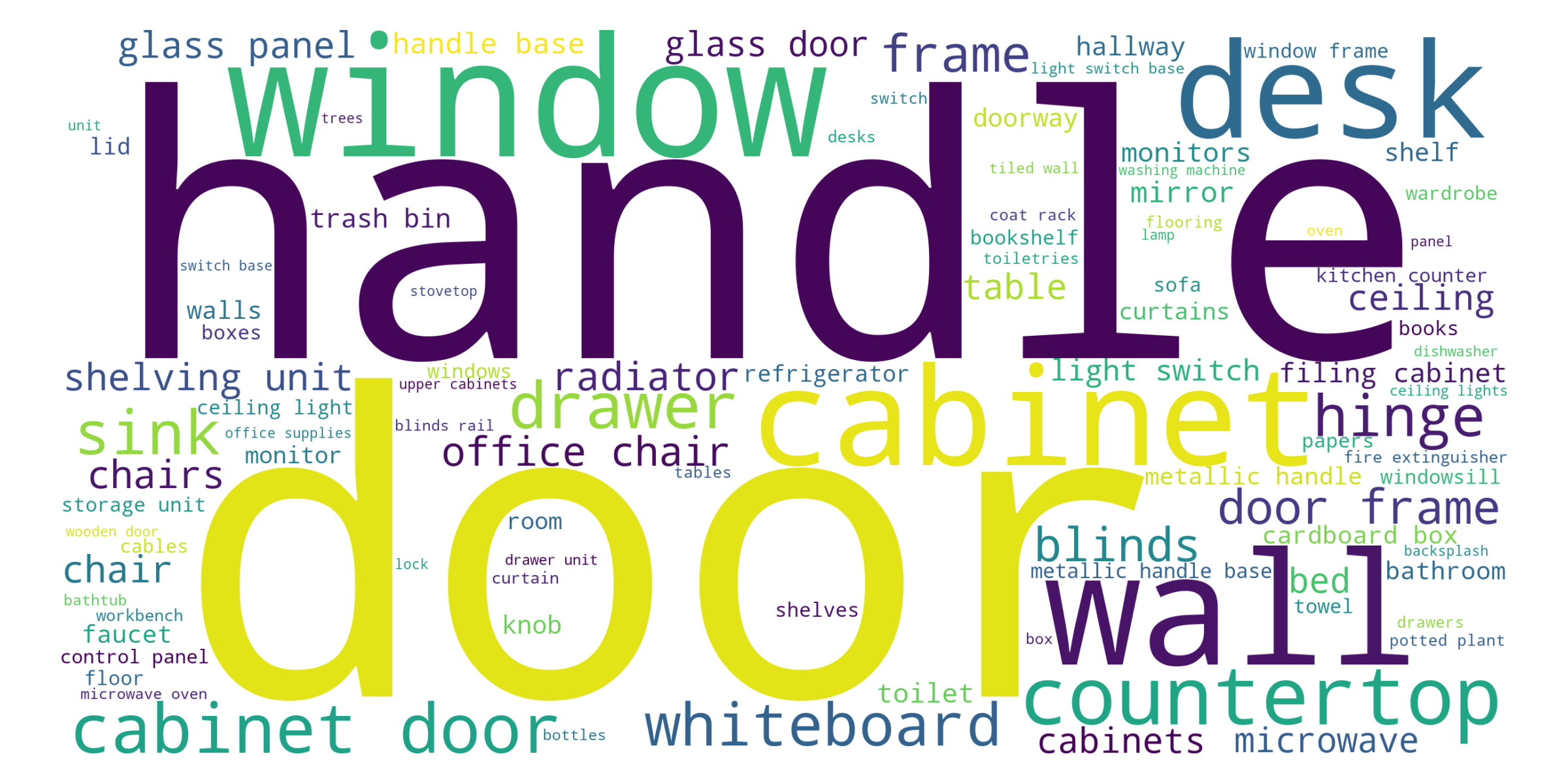}   
        \caption{object names}
        \label{fig:word_clouds_a}
      \end{subfigure}
      \begin{subfigure}{0.49\linewidth}
        \includegraphics[width=\linewidth]{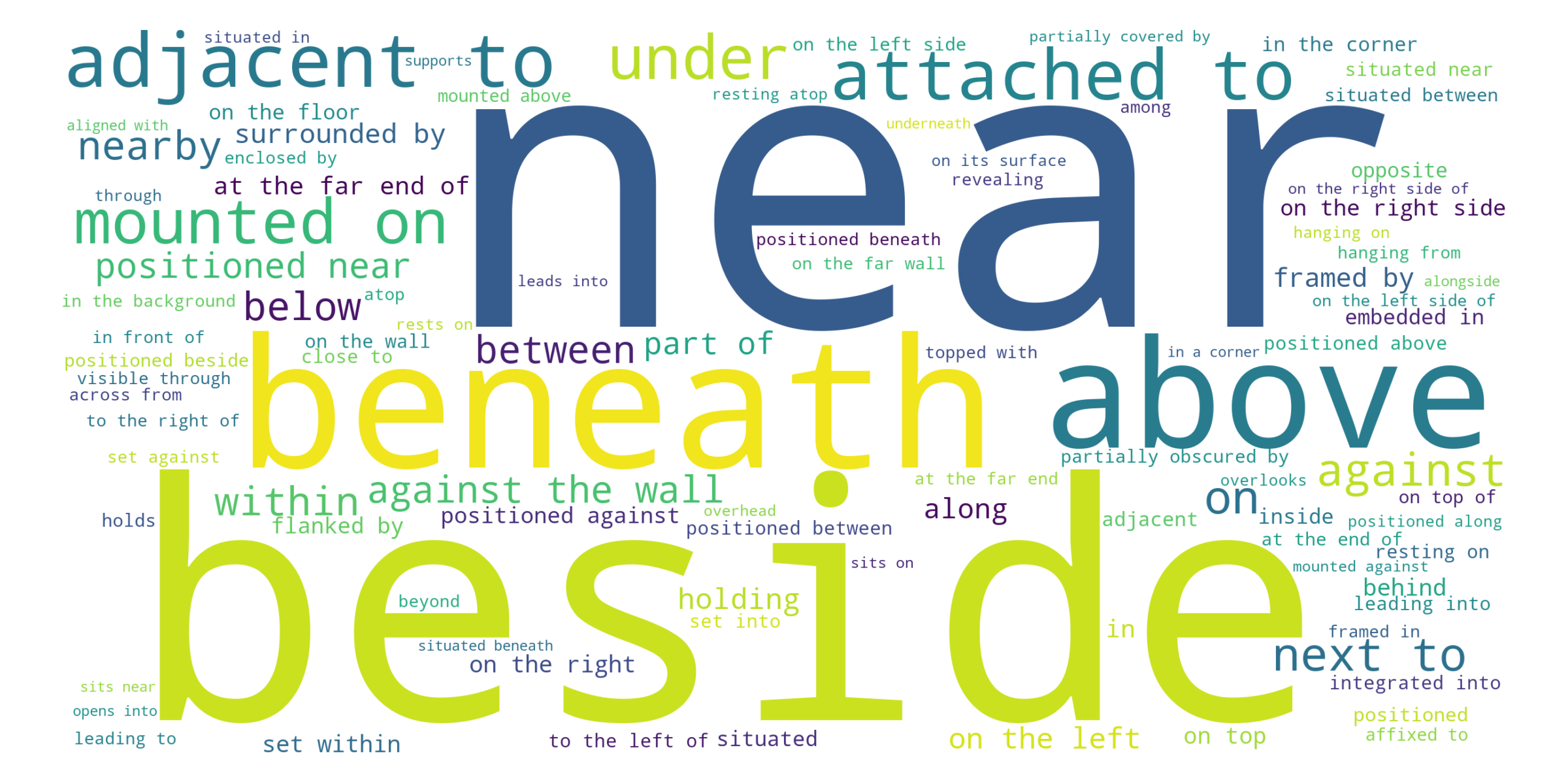}
        \caption{spatial relations}
        \label{fig:word_clouds_b}
      \end{subfigure}
      \begin{subfigure}{0.32\linewidth}
        \includegraphics[width=\linewidth]{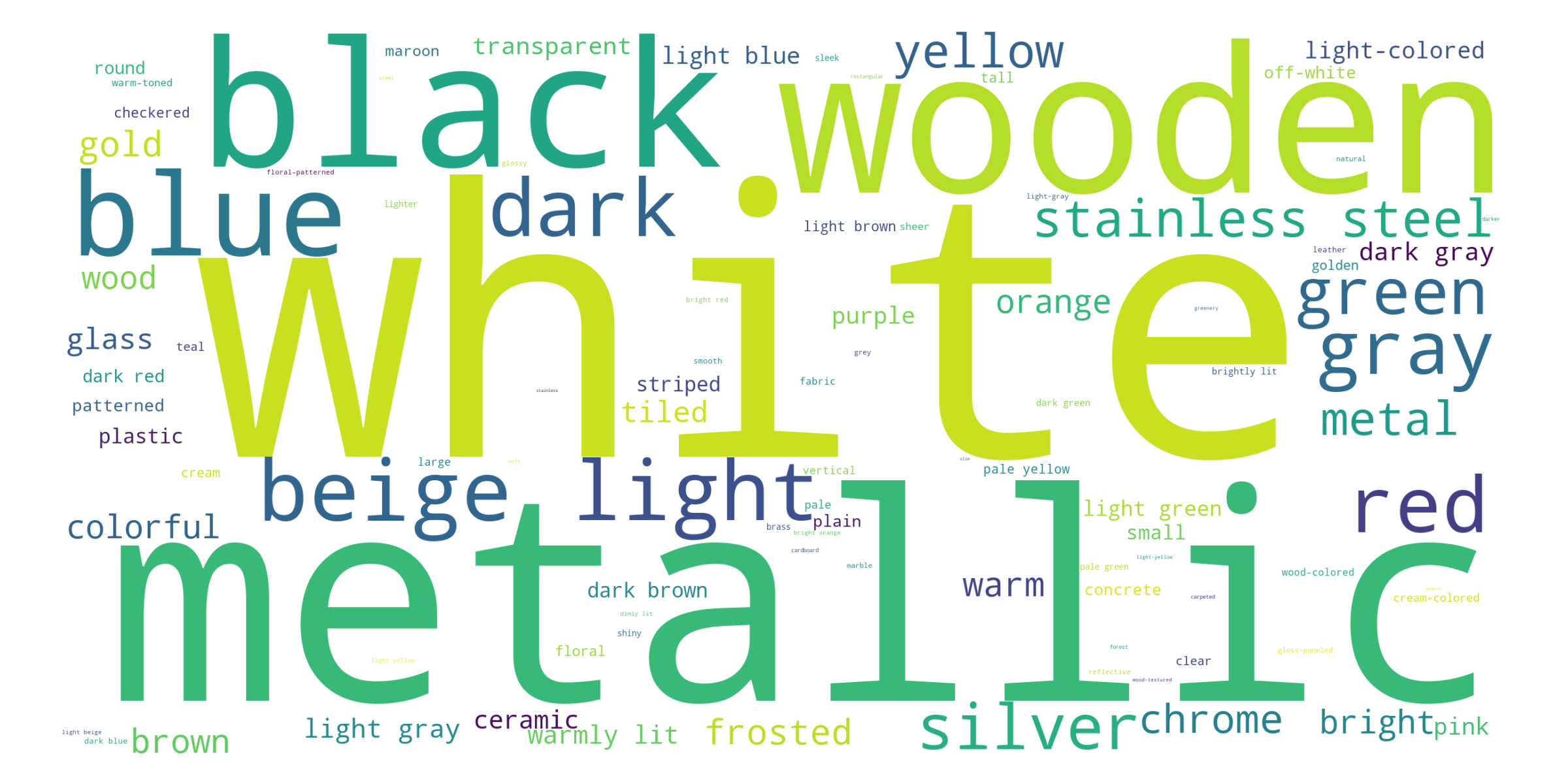}
        \caption{colors}
        \label{fig:word_clouds_c}
      \end{subfigure}
      \begin{subfigure}{0.32\linewidth}
        \includegraphics[width=\linewidth]{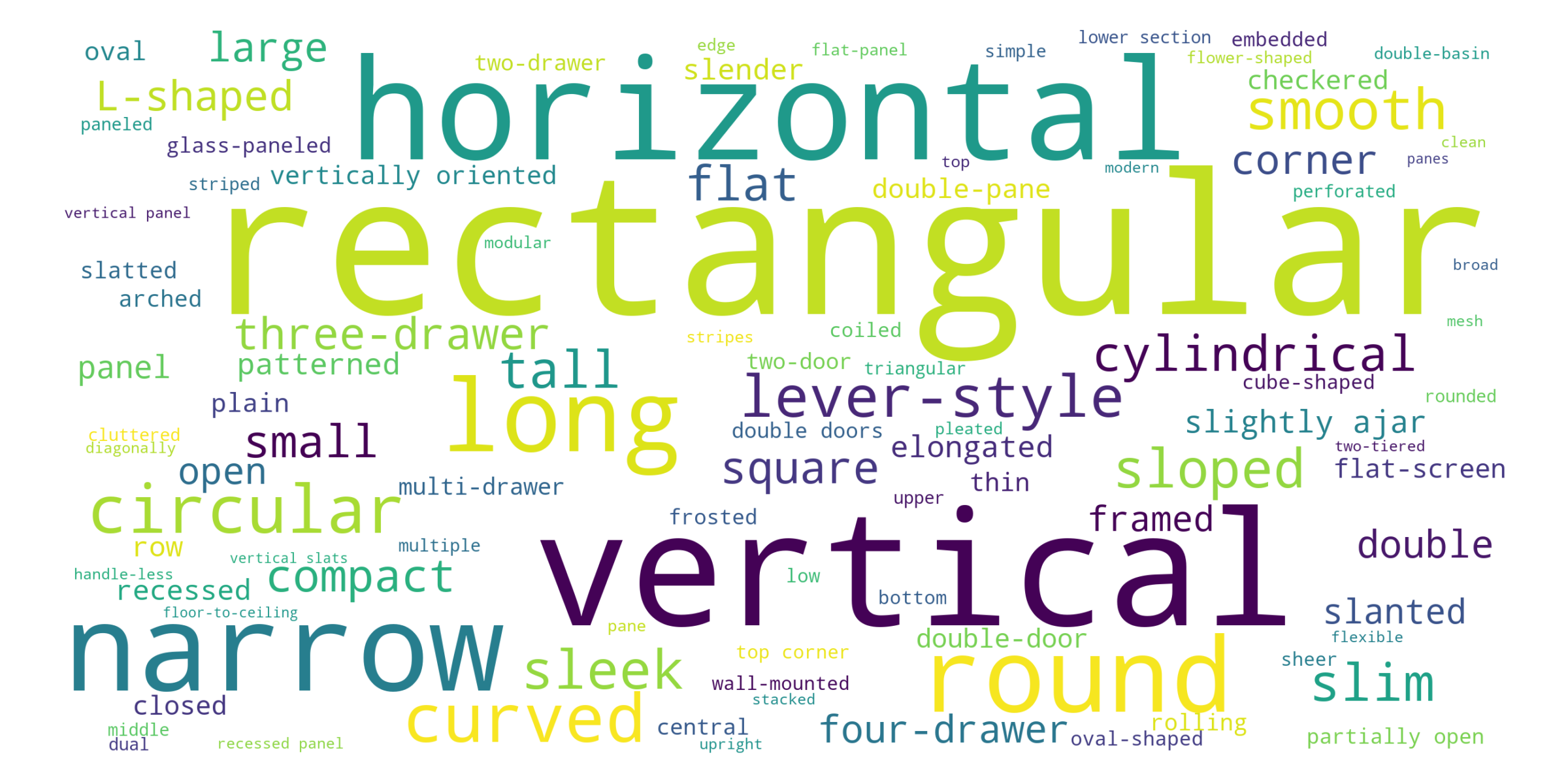}
        \caption{shapes}
        \label{fig:word_clouds_d}
      \end{subfigure}
      \begin{subfigure}{0.32\linewidth}
        \includegraphics[width=\linewidth]{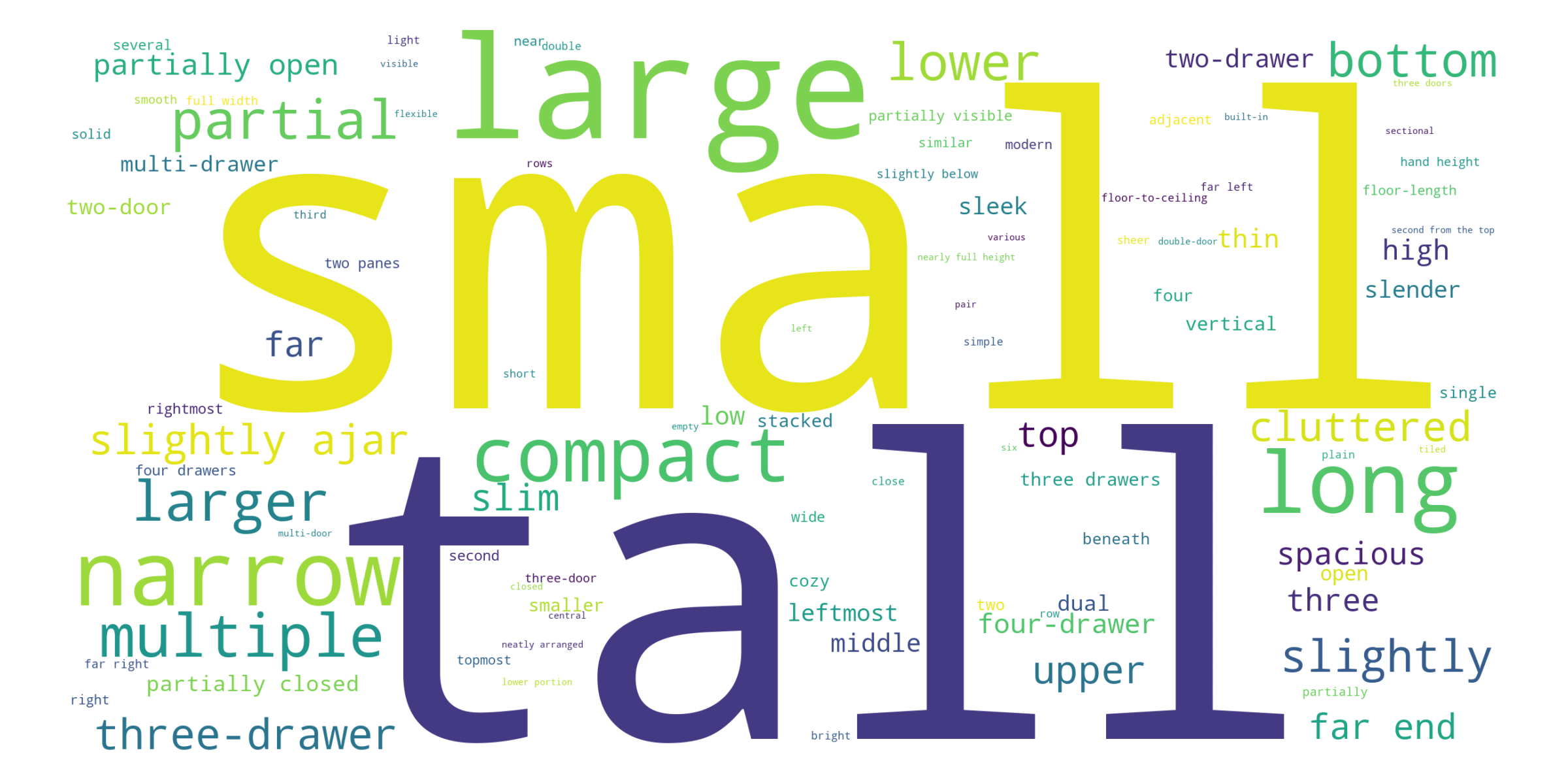}
        \caption{sizes}
        \label{fig:word_clouds_e}
      \end{subfigure}
    
    \caption{\textbf{Dataset Statistics.} (a) shows the word count distribution, indicating balanced conciseness. (b) shows the frequency of sentence categories including attributes (fundamental object properties), parts (object components), intra-class spatial relations (same-category object positions), inter-class spatial relations (different-category object positions), ordinal numbers (sequential order), superlatives (extreme degrees), and comparatives (comparisons), showing syntactic diversity. (c)-(g) are Word clouds of terms for the dataset. Bigger fonts indicate more frequent terms in the descriptions.}
    \label{fig:dataset_statistics}
\end{figure}

\subsection{Dataset Construction Details}

We adopt \textit{Qwen3-VL-235B-A22B-Instruct} \cite{qwen3technicalreport}, with $1.2$ temperature and prompt in \cref{tab:prompt_annotation}, to generate annotations. In multiple VLMs verification, we utilize \textit{Qwen3-VL-235B-A22B-Instruct} \cite{qwen3technicalreport}, \textit{Qwen3-VL-235B-A22B-Thinking} \cite{qwen3technicalreport}, \textit{Qwen2.5-VL-72B-Instruct} \cite{Qwen2.5-VL}, \textit{GLM-4.5V} \cite{vteam2025glm45v}, \textit{Step3} \cite{step3system} and \textit{Deepseek-VL2} \cite{wu2024deepseekvl2}, with same settings that temperature is $0.2$ and prompt is in \cref{tab:prompt_verification}.

\subsection{Dataset Statistics}

OpenTarget starts from 13,194 VLM-generated queries and retains 9,943 queries through multiple VLMs verification, which are then manually reviewed. During manual verification, 2,219 queries are removed and 2,422 are rewritten for better accuracy and clarity, resulting in the final set of \textbf{7,724} valid descriptions with word counts ranging 7-72. As shown in \cref{fig:description_lengths}, the length distribution is roughly normal (mode: 32 words, 5.49\%), with over 98.1\% of descriptions exceeding 16 words, ensuring comprehensive detail. \cref{fig:sentence_distribution} categorizes descriptions into 7 compatible sentence types (a single description can belong to multiple types, so counts sum to more than 7,724, as illustrated in \cref{tab:sentence_type_examples}). Three dominant categories emerge: 'attributes' (7,039, 91.13\%) is fundamental and easily combined with other categories; 'parts' (5,642, 73.04\%) is driven by the dataset's part-level focus; 'inter-class spatial relations' (4,705, 60.91\%) is inflated by part-parent object descriptions. Remaining types (intra-class spatial relations, ordinals, superlatives, comparatives) ensure syntactic diversity. 
Word clouds (\cref{fig:word_clouds_a,fig:word_clouds_b,fig:word_clouds_c,fig:word_clouds_d,fig:word_clouds_e}) demonstrate notable lexical richness: as illustrated, vocabulary related to object names, spatial relations, colors, shapes, and sizes is all diverse and abundant, enabling precise and varied descriptions.

\section{Comparative Analysis and Necessity of $\olt$}

\paragraph{OLT-Free (Necessity of OLT).}
Although the inability to ground objects outside the predefined object lookup table ($\olt$) may appear to be caused by the $\olt$ itself, a straightforward alternative---removing the $\olt$ entirely---is not a practical solution.
In fact, early zero-shot 3DVG methods \cite{yang2024llm,Peng2023OpenScene,conceptfusion} were developed without the notion of an $\olt$.
They were mainly CLIP-based and relied on direct feature matching between textual queries and 3D point clouds.
However, such matching is essentially analogous to retrieving candidates whose features correlate with the queried object label, and its effective granularity is entirely determined by the underlying point-cloud feature model.
As a result, these methods have limited ability to support flexible multi-granularity grounding, especially when the target shifts from coarse categories to fine-grained parts.

More recent \textbf{OLT-Free} method, \eg VLM-Grounder \cite{xu2024vlmgrounder}, avoids this dependence on point-cloud feature representations by introducing 2D foundation models to perceive previously unseen objects, which appears to offer a more general solution.
However, in practice, it remains inefficient even for common objects.
Without the assistance of an $\olt$, grounding in the same scene repeatedly requires searching nearly the entire scene from scratch, since no persistent object-level representation is maintained across queries or sub-goals.
This leads to substantial computational overhead and makes such methods difficult to deploy in realistic settings.
Therefore, at the current stage, the $\olt$ remains necessary: it allows object entries to be defined at flexible granularity, and meanwhile serves as a form of persistent memory that can greatly accelerate grounding.

\paragraph{OLT-Completion v.s. Ours.}
Since the $\olt$ is both necessary and flexible, one may naturally ask whether the problem can be solved simply by constructing a sufficiently complete $\olt$, thereby avoiding the need to ground objects outside it.
However, existing OLT-based methods \cite{li2025seeground, jin2025spazer, wang2025affordbot, lin2025seqvlm} build the table before grounding, i.e., in an offline manner.
Under open-world queries, there will always exist target objects that fall outside such a predefined offline OLT.
For this reason, \textbf{OLT-Completion} is fundamentally an unbounded objective:
strategies such as improving the segmentor or combining multiple segmentors may enlarge the coverage of the $\olt$, but they cannot eliminate the problem at its root.
Moreover, these strategies typically introduce substantial preprocessing cost and enlarge the candidate space for subsequent grounding.

In contrast, our method is query-adaptive rather than exhaustive.
OpenGround starts from an incomplete but efficient offline OLT, and invokes Context-Guided Perception (CGP) only when the current sub-goal cannot be matched to existing entries.
More importantly, this perception process is guided by the task chain and previously grounded contextual objects, so the discovery procedure is restricted to spatially relevant regions instead of the whole scene.
In this way, the problem is reformulated from offline exhaustive completion into online demand-driven completion.
Therefore, compared with \textbf{OLT-Completion}, our method achieves a better balance among coverage, efficiency, and compatibility.
It avoids the impractical goal of constructing a universally complete offline OLT, while preserving the advantages of OLT-based grounding for common objects.

\section{Method Details}

\subsection{Supplementary for Task Chain Construction}

We conduct a user study to quantify whether the task chains generated by our planner resemble the progressive grounding strategies adopted by humans. A total of 41 participants are recruited, and each participant completes 10 query instances, yielding 410 human annotations in total. For each query, participants are asked to provide an object-level ordering that best matches how they would progressively locate the target, starting from readily identifiable contextual objects and proceeding toward the final target object. Based on the user study results, we compute the \textbf{Weighted Average Edit Distance (WAED)} which is widely used to calculate similarity between sequences, to evaluate human-likeness of grounding strategies. All computations follow the formulation below.

\paragraph{Levenshtein Edit Distance.}
Given two sequences $\mathbf{a}$ and $\mathbf{b}$, the Levenshtein edit distance
$\mathrm{ED}(\mathbf{a},\mathbf{b})$ is defined as the \emph{minimum number of edit
operations} required to transform $\mathbf{a}$ into $\mathbf{b}$, where the allowed
operations are: insertion of one element, deletion of one element and substitution of one element.

\paragraph{Model-to-Human WAED.}
For each task instance $t$, the human questionnaire yields a set of sequences 
$\{\mathbf{h}_{t,i}\}$ with corresponding frequencies (weights) $w_{t,i}$.
For a model prediction $\mathbf{s}_t$, the weighted average edit distance is:

\begin{equation}
\mathrm{WAED}(\mathbf{s}_t)
=
\frac{
\sum_{i} 
w_{t,i} \; \mathrm{ED}(\mathbf{s}_t, \mathbf{h}_{t,i})
}{
\sum_i w_{t,i}
}.
\end{equation}
Averaging across all sequences gives the final reported score:

\begin{equation}
\mathrm{WAED}_{\text{model}}
=
\frac{1}{T}
\sum_{t=1}^{T}
\mathrm{WAED}(\mathbf{s}_t).
\end{equation}

\paragraph{Human–Human Internal Disagreement.}
Within each task instance $t$, humans produce multiple valid orderings.  
We measure their internal disagreement by computing the pairwise weighted edit distance:

\begin{equation}
\mathrm{WAED}^{\text{human}}_t
=
\sum_{i < j}
\left(
\frac{w_{t,i} \, w_{t,j}}
     {(\sum_k w_{t,k})^2}
\right)
\mathrm{ED}(\mathbf{h}_{t,i}, \mathbf{h}_{t,j}).
\end{equation}
Finally, the overall human–human inconsistency level is:

\begin{equation}
\mathrm{WAED}_{\text{human}}
=
\frac{
\sum_{t} 
\left( 
(\sum_k w_{t,k}) \, \mathrm{WAED}^{\text{human}}_t 
\right)
}{
\sum_{t} \sum_k w_{t,k}
}.
\end{equation}
This is the internal disagreement level between human annotators and serves as the
reference baseline in \cref{fig:task_chain_construction_process}.

\subsection{Context-Guided Perspective Selection}

We show the pseudo code of \cref{sec:ace} in \cref{alg:perspective_selection_ace}, including fallback strategy for non previously grounded objects provided. The fallback strategy iteratively samples views whose observations contribute sufficiently new, non-overlapping 3D points to the accumulated global scene coverage. Each candidate view is evaluated by the proportion of newly observed points, and only views exceeding a minimal coverage threshold are retained. This ensures efficient scene exploration while avoiding redundant observations.

\paragraph{Registered Candidate Views.}
Candidate observations are registered RGB-D frames from the input sequence, rather than virtual views rendered from resampled camera poses. Following the frame-sampling protocol used by prior grounding systems, we uniformly retain one frame every 20 frames. This results in approximately 56 candidate views per scene on average.

\paragraph{Visible Points Calculation.} Let $\mathcal{P}$ denote the global scene point cloud. For each candidate view $k$, we precompute its visible-point set $\mathcal{P}_k$. Let
\begin{equation}
\Pi_k(\mathbf{p})=\bigl(u_k(\mathbf{p}),v_k(\mathbf{p}),z_k(\mathbf{p})\bigr)
\end{equation}
denote the projection of a global 3D point $\mathbf{p}$ into view $k$, where $z_k(\mathbf{p})$ is its depth in the corresponding camera coordinate system. Given the registered depth map $D_k$, the visible-point set is
\begin{equation}
\mathcal{P}_k=\left\{
\mathbf{p}\in\mathcal{P}\ \middle|\
\begin{array}{l}
(u_k(\mathbf{p}),v_k(\mathbf{p}))\in\Omega_k,\\
D_k(u_k(\mathbf{p}),v_k(\mathbf{p}))>0,\\
z_k(\mathbf{p})-D_k(u_k(\mathbf{p}),v_k(\mathbf{p}))<\epsilon
\end{array}
\right\},
\end{equation}
where $\Omega_k$ denotes the image domain and $\epsilon$ is a depth-consistency tolerance. Visibility is therefore computed only once. Subsequent view selection requires only set operations, without repeated projection or rendering.

\paragraph{From 2D Masks to 3D Proposals.}
After perspective selection, CGP applies 2D segmentation to the selected observations. Let $M_{k,m}$ denote the $m$-th valid mask predicted in view $k$. For every valid depth pixel $(u,v)\in M_{k,m}$, its camera-coordinate 3D point is obtained by
\begin{equation}
\mathbf{x}^{c}_{k,m}(u,v)=D_k(u,v)\,\mathbf{K}_k^{-1}
\begin{bmatrix}
u & v & 1
\end{bmatrix}^{\top},
\end{equation}
where $\mathbf{K}_k$ is the camera intrinsic matrix. The point is then transformed into the global coordinate system using the registered camera pose:
\begin{equation}
\mathbf{x}^{w}_{k,m}(u,v)=\mathbf{T}_{w\leftarrow k}\mathbf{x}^{c}_{k,m}(u,v).
\end{equation}
The back-projected points associated with one mask form a lifted 3D proposal. We compute a 3D bounding box for each proposal and iteratively merge proposals whose boxes satisfy
\begin{equation}
\mathrm{IoU}_{3\mathrm{D}}(B_a,B_b)>\tau_{\mathrm{iou}},\qquad\tau_{\mathrm{iou}}=0.5.
\end{equation}
Merging continues until no remaining proposal pair exceeds the threshold. The resulting instances are assigned new object identifiers and inserted into the online $\olt$, after which they can be processed by the same Single-Step Grounding module as existing objects.

\begin{algorithm}
\caption{Context-Guided Perspective Selection}
\label{alg:perspective_selection_ace}
\begin{algorithmic}[1]
\Require Previously grounded objects $O=\{O_1,\dots,O_{t-1}\}$, point clouds $\pcd$, max views $V$, all RGB images $\rgbset$
\State Initialize observed regions $\mathcal{R}_{O_i} \gets \emptyset$ for all $O_i$
\State $\rgbset^\star \gets \emptyset$
\For{$v = 1$ to $V$}
    \State $\Delta^\star \gets 0$, $k^\star \gets \text{None}$
    \For{each candidate view $k \notin \rgbset^\star$}
        \State $\Delta \gets 0$
        \For{each object $O_i$}
            \State $new\_pts \gets (\pcd_{O_i} \cap \pcd_k) \setminus \mathcal{R}_{O_i}$
            \State $\Delta \gets \Delta + |new\_pts|/|\pcd_{O_i}|$
        \EndFor
        \If{$\Delta > \Delta^\star$}
            \State $\Delta^\star \gets \Delta$
            \State $k^\star \gets k$
        \EndIf
    \EndFor

    \If{$k^\star = \text{None}$}
        \State \textbf{break}
    \EndIf

    \State $\rgbset^\star \gets \rgbset^\star \cup \{\rgb_{k^\star}\}$

    \For{each object $O_i$}
        \State $\mathcal{R}_{O_i} \gets \mathcal{R}_{O_i} \cup (\pcd_{O_i} \cap \pcd_{k^\star})$
    \EndFor

    \If{$\forall O_i,\ \mathcal{R}_{O_i} = \pcd_{O_i}$}
        \State \textbf{break}
    \EndIf
\EndFor

\If{$\rgbset^\star=\emptyset$} \Comment{Fallback strategy: observe whole scene}
    \State $\mathcal{OP} \gets \emptyset$
    \State Define threshold $\tau$
    \For{each candidate view $k$}
        \State $new\_pts \gets \pcd_k \setminus \mathcal{OP}$
        \If{$\frac{|new\_pts|}{|\pcd_k|} < \tau$}
            \State \textbf{continue}
        \EndIf
        \State $\rgbset^\star \gets \rgbset^\star \cup \{\rgb_k\}$
        \State $\mathcal{OP} \gets \mathcal{OP} \cup \pcd_k$
    \EndFor
\EndIf
\State \Return $\rgbset^\star$
\end{algorithmic}
\end{algorithm}

\subsection{Single-Step Grounding}
\label{sec:grounding}

As illustrated in \cref{tab:ablation_comparison}, our single-step grounding backbone outperforms other methods, \eg SeeGround \cite{li2025seeground} and SeqVLM \cite{lin2025seqvlm}. While the strong backbone contributes to the final performance, it is not the central contribution of our work, thus it is organized in the supplementary material. Within OpenGround, the Single-Step Grounding operates on the updated online $\olt$ and additionally leverages previously grounded objects as context guidance. This contextual information is exploited in perspective selection, annotation construction, and downstream VLM reasoning, which together make candidate verification more robust under complex queries and cluttered scenes.  
We introduce details below.

\paragraph{Perspective Selection.} 
As illustrated in \cref{alg:perspective_selection_ssg}, we further incorporate previously grounded objects from the task chain as context guidance in Single-Step Grounding. Given a candidate object $c_j$ from all candidates retrieved from the current online $\olt$, we first apply \textsc{Context-Guided Perspective Selection} (in \cref{alg:perspective_selection_ace}) over previously grounded objects $O=\parB{O_1,\cdots,O_{t-1}}$ to obtain a compact set of context-relevant views, denoted by $\rgbset^\star$.
This step identifies the views that best preserve the already established context guidance, and therefore serves as a coarse filtering stage before candidate verification.
If the candidate object $c_j$ does not appear in $\rgbset^\star$, we directly reject it and return an empty result, since such a candidate is unlikely to be consistent with the grounded context.\
If $c_j$ remains visible in the filtered view set, we further concatenate it to the grounded object set as the current object $O_t$, and perform a second greedy view selection procedure.

In this way, context guidance is used not only to support missing-object discovery in CGP, but also to improve standard single-step grounding on the online $\olt$.
Compared with directly selecting views for each candidate independently, this design encourages the VLM to verify the candidate together with the contextual evidence accumulated along the task chain, while also filtering out context-inconsistent candidates at an early stage. As illustrated in \cref{tab:ssg_ab_study}, our Context-Guided Perspective Selection boost Single-Step Grounding.

\begin{algorithm}
\caption{Context-Guided Perspective Selection for Single-Step Grounding}
\label{alg:perspective_selection_ssg}
\begin{algorithmic}[1]
\Require Candidate object $c_j$, previously grounded objects $O=\parB{O_1,\cdots,O_{t-1}}$, point clouds $\pcd$, max views $V$, all RGB images $\rgbset$

\State $\rgbset^\star \gets \text{Context-Guided Perspective Selection}(O,\pcd,V,\rgbset)$
\If{$c_j$ does not exist in $\rgbset^\star$}
    \State \Return $\emptyset$
\EndIf

\State Concatenate $c_j$ to $O$ as $O_t$ \Comment{See $c_j$ as Context Guidance}
\State Initialize observed regions $\mathcal{R}_{O_i}\gets \emptyset$ for all $O_i$
\State $\mathcal{R}_{c_j} \gets \emptyset$, $\rgbset^\star_j \gets \emptyset$
\For{$v = 1$ to $V$}
    \State $\Delta^\star \gets 0$, $k^\star \gets \text{None}$
    \For{each candidate view $k \notin \rgbset^\star_j$}
        \State $\Delta \gets 0$
        \For{each object $O_i$}
            \State $new\_pts \gets (\pcd_{O_i} \cap \pcd_k) \setminus \mathcal{R}_{O_i}$
            \State $\Delta \gets \Delta + |new\_pts|/|\pcd_{O_i}|$
        \EndFor
        \If{$\Delta > \Delta^\star$}
            \State $\Delta^\star \gets \Delta$
            \State $k^\star \gets k$
        \EndIf
    \EndFor

    \If{$k^\star = \text{None}$}
        \State \textbf{break}
    \EndIf

    \State $\rgbset^\star_j \gets \rgbset^\star_j \cup \{\rgb_{k^\star}\}$

    \For{each object $O_i$}
        \State $\mathcal{R}_{O_i} \gets \mathcal{R}_{O_i} \cup (\pcd_{O_i} \cap \pcd_{k^\star})$
    \EndFor

    \If{$\forall O_i,\ \mathcal{R}_{O_i} = \pcd_{O_i}$}
        \State \textbf{break}
    \EndIf
\EndFor

\State \Return $\rgbset^\star_j$
\end{algorithmic}
\end{algorithm}

\begin{figure}[t]
    \centering
    \begin{subfigure}{0.49\linewidth}
        \includegraphics[width=\linewidth]{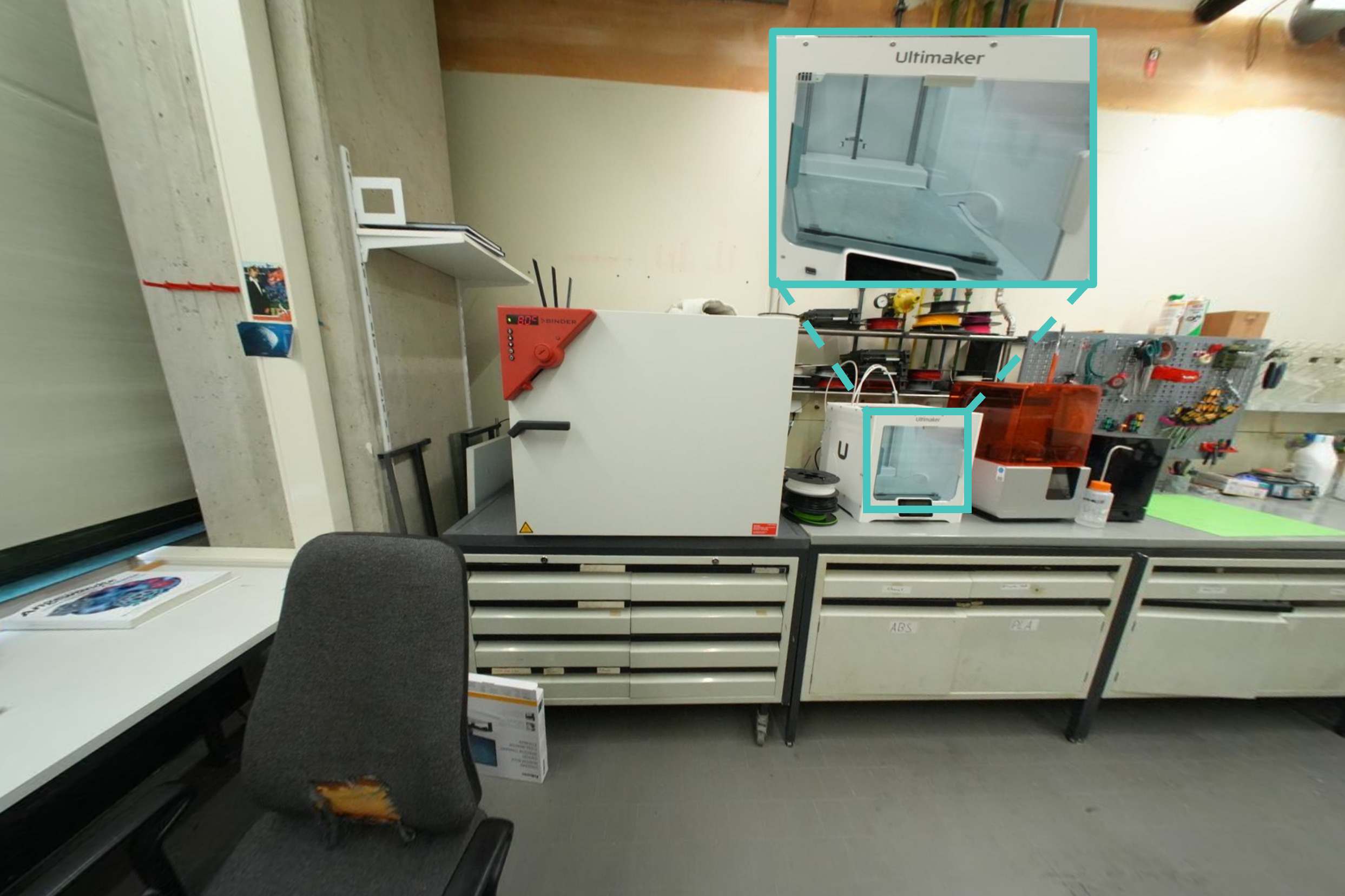}
        \caption{Origin image.}
        \label{fig:annotation_comp_a}
    \end{subfigure}
    \begin{subfigure}{0.49\linewidth}
        \includegraphics[width=\linewidth]{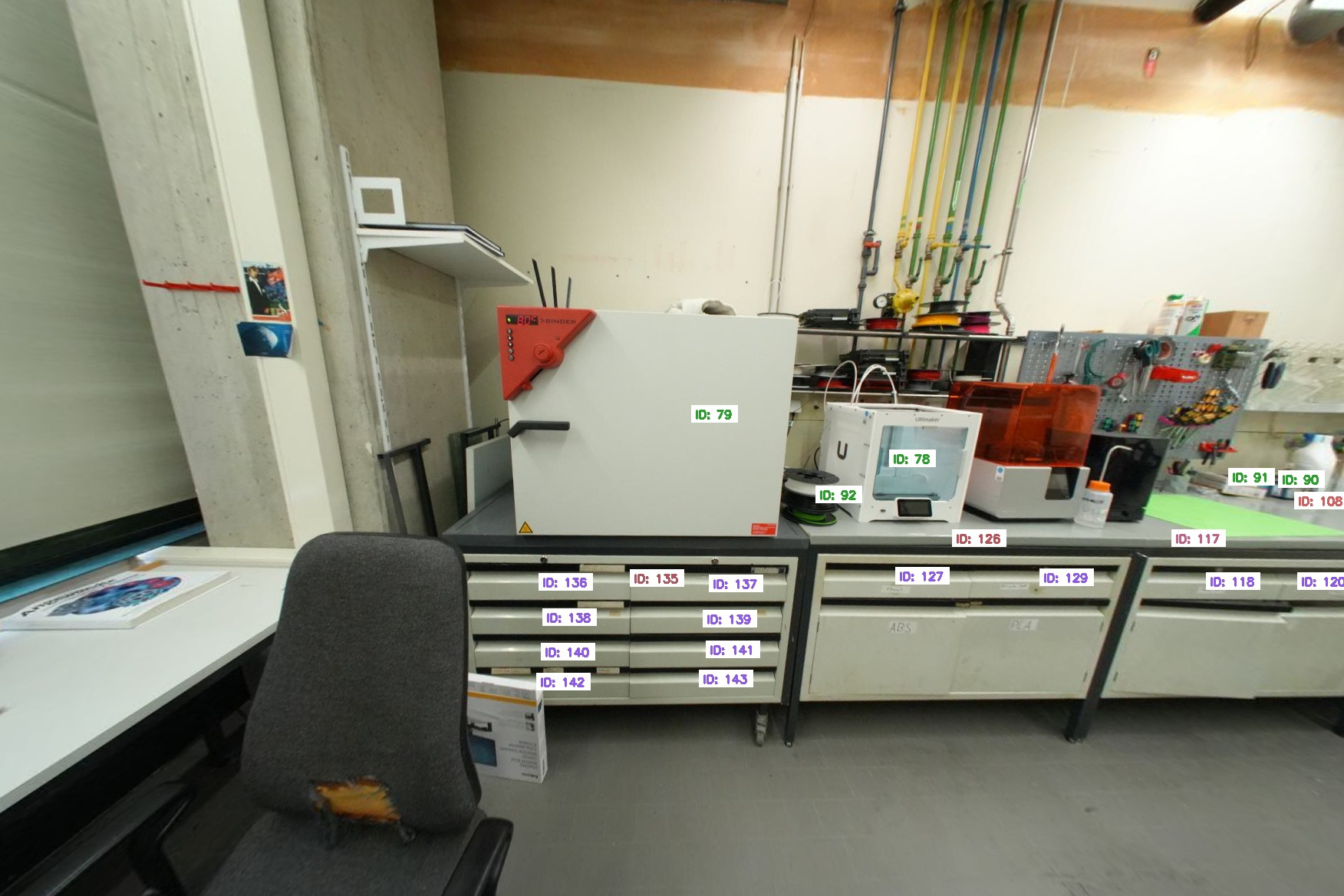}
        \caption{All objects mentioned.}
        \label{fig:annotation_comp_b}
    \end{subfigure}
    \begin{subfigure}{0.49\linewidth}
        \includegraphics[width=\linewidth]{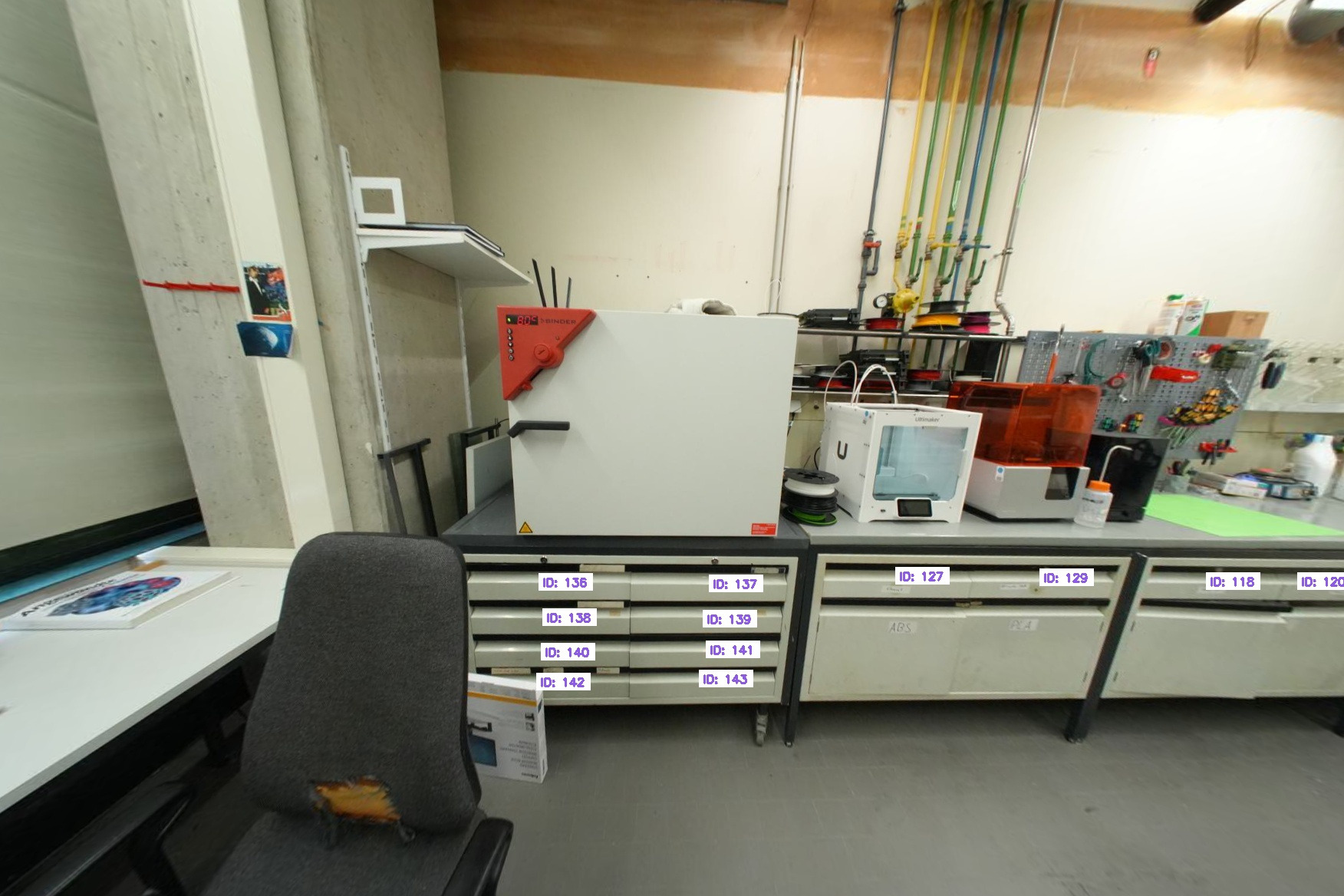}
        \caption{All candidates.}
        \label{fig:annotation_comp_c}
    \end{subfigure}
    \begin{subfigure}{0.49\linewidth}
        \includegraphics[width=\linewidth]{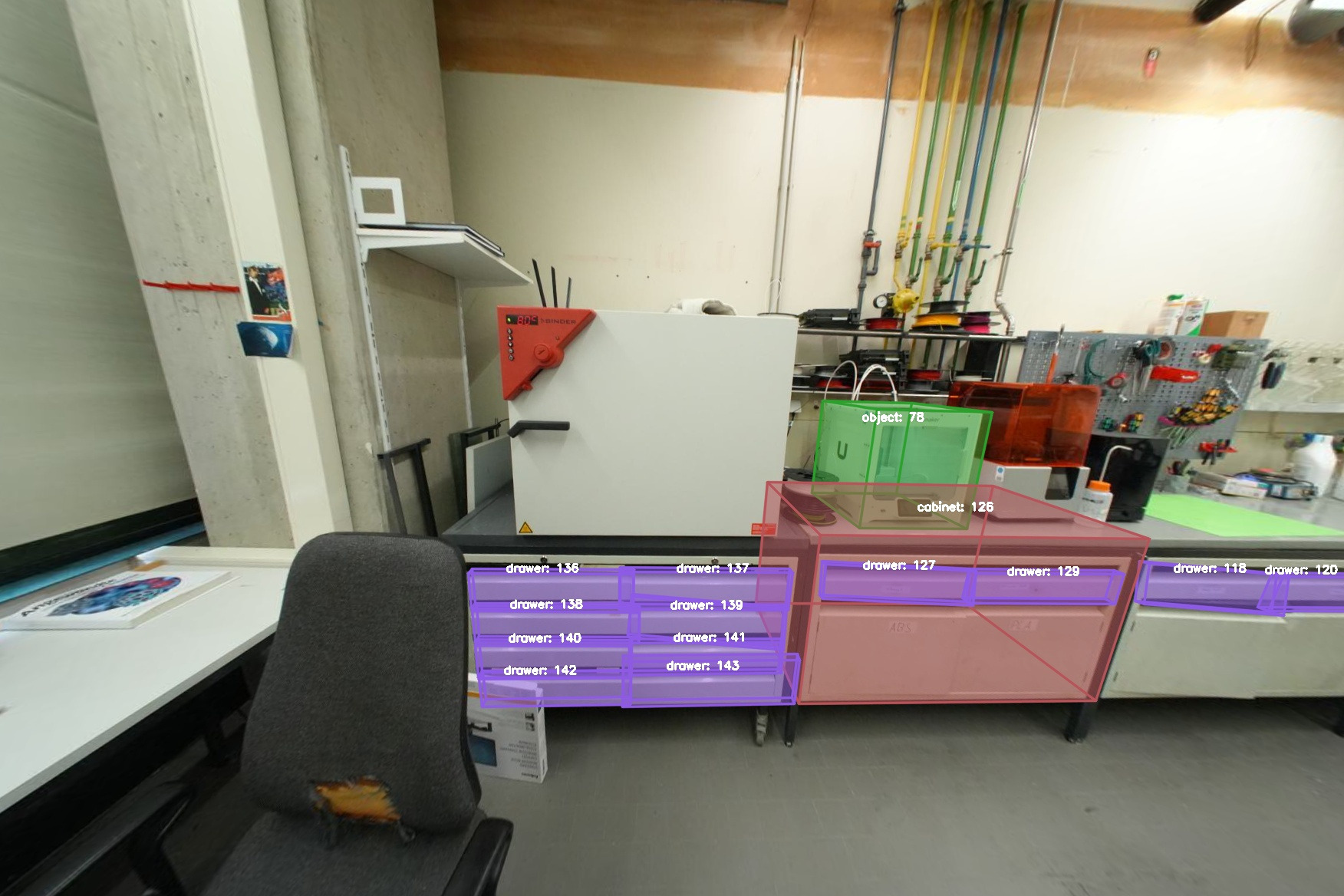}
        \caption{Ours}
        \label{fig:annotation_comp_d}
    \end{subfigure}
    \caption{\textbf{Annotation Visualization.} The query is ``\textit{Locate the drawer directly beneath the machine-like object marked with `Ultimaker'. The drawer is a part of the cabinet under the machine-like object.}'' (a) is the origin image without any annotations. (b) annotates all objects mentioned (``drawer'', ``cabinet'', ``object'') in the query. (c) only annotates all candidates for ``drawer''. (d) annotates previously grounded objects and all candidates. }
    \label{fig:annotation_comp}
\end{figure}

\paragraph{Annotation.} As a supplementary explanation, \cref{tab:ssg_ab_study} illustrates the impact of different annotation methods, revealing substantial performance differences. Here, as illustrated in \cref{fig:annotation_comp}, we provide a qualitative visualization of representative samples to intuitively demonstrate how these annotation strategies lead to noticeably different results. As shown in \cref{fig:annotation_comp}, our proposed annotation strategy effectively balances informativeness and distraction, leading to more accurate and robust grounding.

\begin{itemize}
    \item \textbf{All Objects Mentioned.} It annotates all query-relevant objects introduces excessive distractions (\textit{e.g.}, irrelevant ``object'' which is green in \cref{fig:annotation_comp_b}), overwhelming the VLM with irrelevant regions. It discards partial context guidance from task chain.
    \item \textbf{All Candidates.} It annotates only candidates, missing critical spatial context from previously grounded objects (\textit{e.g.}, ``the machine-like object marked with `Ultimaker''' which is only visible when close enough). It discards partial context guidance from task chain.
    \item \textbf{Ours.} Our method marks previously grounded objects and current candidates. Since previously grounded objects are definitively localized, each contributes exactly a single annotation (\textit{e.g.}, ``object:78'' and ``cabinet:126''). This avoids the distraction from previously grounded objects in strategy (b) while preserving critical context from grounded objects, which strategy (c) fails to provide.
\end{itemize}

\paragraph{VLM Reasoning.} Inspired by prior works \cite{huang2025viewsrd, xu2024vlmgrounder, lin2025seqvlm}, we feed the query $\query$ to the VLM to extract structured semantic conditions for the target label $L_{T_t}$. The structured decomposition disentangles the complex linguistic instructions into interpretable criteria, such as spatial relations and attributes, facilitating precise reasoning in the subsequent stage. Next, both annotated $\rgbset_A^\star$ and non-annotated $\rgbset^\star$ are provided to the VLM for multi-view reasoning. In particular, annotated views supply explicit spatial cues of objects, while non-annotated ones preserve holistic visual perception without potential visual occlusion caused by excessive annotations, as observed in SeeGround \cite{li2025seeground}. Then, VLM checks which object satisfies all conditions and outputs the grounded object $O_t$ for task $T_t$. As illustrated in \cref{tab:ssg_ab_study}, our VLM Reasoning also boosts Single-Step Grounding.

\begin{table}
    \centering
    \caption{Ablation Study on Different Components of Single-Step Grounding.}
    \scalebox{0.8}{
    \begin{tabular}{l|c|c|c|c}
        \toprule
        \# & Perspective Selection & Annotation & VLM Reasoning & Acc@0.50 \\
        \midrule
        \midrule
        (1) & Ours & Ours & Ours & 34.2 \\
        \midrule
        (2) & \textit{SeeGround} \cite{li2025seeground} & Ours & Ours & 31.6\\
        (3) & \textit{SeqVLM} \cite{lin2025seqvlm} & Ours & Ours & 31.8\\
        (4) & \textit{Struct2D} \cite{zhu2025struct2d} & Ours & Ours & 26.3\\
        \midrule
        (5) & Ours & \textit{All Objects Mentioned} & Ours & 25.3\\
        (6) & Ours & \textit{All Candidates} & Ours & 26.5\\
        \midrule
        (7) & Ours & Ours & \textit{Non Decomposition} & 32.9\\
        (8) & Ours & Ours & \textit{Non Origin Images} & 33.6\\
        \bottomrule
    \end{tabular}
    }
    \label{tab:ssg_ab_study}
\end{table}

\section{Implementation Details}

For the GroundedSAM \cite{ren2024grounded}, we utilize Swin-B GroundingDINO Decoder \cite{liu2023grounding} with text and box thresholds both set to 0.4 , and utilize ViT-H SAM2 \cite{ravi2024sam2} to generate instance masks from the bounding boxes generated by GroundingDINO, which is also applied to VLM-Grounder \cite{xu2024vlmgrounder} on OpenTarget. We set VLM temperature to 0.2 for all steps in OpenGround framework and other methods.

\subsection{Details of Prompt Designs}

We adopts different prompts for different tasks of the VLM. We conduct prompt example in \cref{tab:prompt_object_parsing} for \textbf{Objects Parsing} in \cref{sec:task_chain_construction}, \cref{tab:prompt_task_chain_construction} for \textbf{Task Chain Construction} in \cref{sec:task_chain_construction}, \cref{tab:prompt_conditions_retrieval} for conditions extraction in \textbf{VLM Reasoning}, and \cref{tab:prompt_vlm_reasoning} for reasoning in \cref{sec:grounding}.

\subsection{Details of Compared Methods}

For a fair and rigorous comparison, we reproduce recently top-performing and open-source baselines on OpenTarget under a unified evaluation protocol, using same VLM (GLM-4.5V \cite{vteam2025glm45v}) for methods requiring VLM. Below we describe our reproduction and adaptations necessary to run each method on our benchmark.

\paragraph{SeeGround \cite{li2025seeground} + GT.} 
Since SeeGround is designed to consume ScanRefer-style inputs, we reorganize our OpenTarget annotations into the ScanRefer format. The ground-truth $\olt$ (object lookup table) is treated as the predicted object proposals required by SeeGround, enabling it to bypass its default grounding step. All other components strictly follow the official implementation.

\paragraph{SPAZER \cite{jin2025spazer} + GT.}
For SPAZER, we use steps described for ``SeeGround + GT'' and follow the settings in its original paper: the number of views $n$ is set to 4, and the Top-$k$ parameter is set to $k$ = 4. 

\paragraph{SeqVLM \cite{lin2025seqvlm} + GT.} 
We feed the ground-truth $\olt$ as its $\mathcal{OPT}$. Following the settings in the original paper, we generate a multi-view sequence with ($n_{\text{frame}}=5$) for each candidate object, and we apply the same VLM-based reasoning procedure with a chain length of ($L = 4$) in official implementation. The remainder, including the iterative reasoning logic, remains unchanged.

\paragraph{VLM-Grounder \cite{xu2024vlmgrounder}.}
To reproduce VLM-Grounder on our dataset, we uniformly sample the image sequence with a ratio of \(20{:}1\), consistent with the original paper. We adopt all hyperparameter settings as reported. Due to computational considerations, we use GroundingDINO \cite{liu2023grounding} instead of GroundingDINO-v1.5 \cite{ren2024grounding} as the 2D detector.

\paragraph{VLM-Grounder \cite{xu2024vlmgrounder} + GT.}
We further evaluate an oracle variant of VLM-Grounder using ground-truth object information. Building upon the setup above, we replace the View Pre-selection stage with ground-truth projections obtained from $\olt$ to directly generate instance masks. We disable the OV-Detection module and instead use the ground-truth $\olt$ entries whose categories match the target class. The final Multi-View Ensemble Projection step is replaced by directly selecting the most relevant ground-truth bounding box from $\olt$. The above adaption fully assess the upper bound of VLM-Grounder through fully leveraging the ground-truth $\olt$.

\paragraph{GPT4Scene \cite{qi2025gpt4scene} + GT.}
For GPT4Scene, we use their released finetuned model. Following the original pipeline, we uniformly sample 64 images and render one ceiling-free top-down BEV map from the point cloud. We annotate these images using the ground-truth $\olt$. The reasoning procedure then strictly follows the official prompts described in the paper.

\section{Additional Quantitative Results}

\subsection{Supplementary Results of Supervised Methods on OpenTarget}

We report the results of supervised 3DVG methods on OpenTarget in the supplementary material, as they are omitted from the main paper due to space limitations.
We include them here for completeness, although their performance under the OpenTarget setting is consistently weak.

We believe this degradation mainly stems from the limitation of supervised training data.
Most existing supervised methods are trained on closed-set benchmarks with predefined object categories and annotation granularity, whereas OpenTarget explicitly evaluates open-world grounding on fine-grained targets beyond the training datasets.
Consequently, the learned grounding priors of these methods generalize poorly to OpenTarget.
This result further supports the motivation of our work: strong performance on conventional supervised benchmarks does not directly translate to open-world 3DVG, where query-adaptive reasoning and online object discovery are necessary.
The comparison is shown in \cref{tab:opentarget_supervised}.

\begin{table}
    \centering
    \caption{Supervised Methods' Performance on \textbf{OpenTarget}.}
    \scalebox{0.75}{
        \begin{tabular}{l|c|c|cc}
            \toprule
            \multirow{2}{*}{Method} & \multirow{2}{*}{$\olt$} & \multirow{2}{*}{Supervision}&\multicolumn{2}{c}{Overall} \\
            \cmidrule{4-5} 
            &  &  & Acc@0.25 & Acc@0.50 \\
        \midrule
        \midrule
        3D-R1 \cite{huang20253d} & - & Supervised & 1.6 & 0.3 \\
        GS-Reasoner \cite{chen2025gsreasoner} & - & Supervised & 12.1 & 9.7 \\
        TSP3D \cite{guo2025text} & - & Supervised & 1.3 & 0.2 \\
        \midrule
        \textbf{Ours} & Mask3D \cite{schult2022mask3d} & Zero-Shot & \textbf{46.2} & \textbf{34.2} \\
        \bottomrule
    \end{tabular}
    }
    \label{tab:opentarget_supervised}
\end{table}

\subsection{Results on Different Segmentation Models}

\begin{table}
    \centering
    \caption{\textbf{Comparison on different segmentation backbones.} Larger SAM variants yield stronger 2D performance, and integrating open-vocabulary 3D segmentation offers competitive results.}
    \label{tab:segmentation_models}
    \begin{tabular}{c|c|c}
    \toprule
        Model & Type & Acc@0.50 \\
        \midrule
        SAM-B & 2D & 26.7 \\
        SAM-L & 2D & 31.5 \\
        SAM-H(Ours) & 2D & \textbf{34.2} \\
        \midrule
        Open3DIS \cite{nguyen2024open3dis} & 3D & \uline{33.6} \\
    \bottomrule
    \end{tabular}
    \vspace{-8pt}
\end{table}

Since GroundedSAM \cite{ren2024grounded} is one of the most widely adopted and general-purpose open-vocabulary 2D segmentation frameworks, we conduct our 2D-based comparisons by varying only the underlying SAM model size (e.g., ViT–B/L/H), while keeping all other components of our pipeline unchanged. This allows us to isolate the effect of segmentation quality on the downstream grounding performance and ensures that improvements are not confounded by differences in detection or reasoning modules.
In addition to the 2D setting, we also evaluate our method with open-vocabulary 3D segmentation model. For this variant, we replace the original \textbf{2D Segmentation and Lifting} stage with a \textbf{3D Segmentation and Filtering} procedure. Specifically, we directly segment instances in 3D space and then filter out objects that are not visible under the selected views, ensuring consistency with the view-dependent grounding pipeline. 
Results in \cref{tab:segmentation_models} demonstrate that while stronger SAM variants improve performance in 2D-based settings, incorporating open-vocabulary 3D segmentation also provides an competitive results.

\subsection{Supplementary for Comparison with SPAZER}

In the main paper \cref{tab:nr3d_comparison}, we observe that SPAZER \cite{jin2025spazer} performs much better in the ``easy'' category on Nr3D \cite{achlioptas2020referit3d}, leading to its overall advantage on the benchmark. We conjecture that its gain stems from an aggressive candidate selection strategy that removes the single distractor typically present in this split, greatly simplifying the search space. To test whether this heuristic also benefits our pipeline, we integrate SPAZER as Single-Step Grounding module into our framework. As illustrated in \cref{tab:nr3d_comparison}, the improvement on Nr3D \cite{achlioptas2020referit3d} is marginal. Notably, this experiment highlights the compatibility and extensibility of our approach: our framework can seamlessly incorporate external Single-Step Grounding methods, while maintaining strong performance.

In addition, we supplement the OpenTarget results of SPAZER here, which are omitted from the main paper due to space constraints. The completed OpenTarget comparison is reported in \cref{tab:opentarget_comparison_add}. Our method outperforms ``SPAZER'' naturally, benefiting from our Task-Chain Planning and Context-Guided Perception, and also outperforms ``Ours+SPAZER'', benefiting from our Context-Guided Perspective Selection in Single-Step Grounding.

\begin{table}
    \centering
    \caption{Performance on \textbf{OpenTarget}. $^*$ denotes results on randomly selected 300 samples due to low efficiency. Methods fail due to missing objects in the Mask3D $\olt$. \textbf{Easy} contains queries with hierarchy length $\leq 2$ (5,737 samples), while \textbf{Hard} contains queries with hierarchy length $> 2$ (1,987 samples).}
    \scalebox{0.75}{
        \begin{tabular}{l|c|cc|cc|cc}
            \toprule
            \multirow{2}{*}{Method} & \multirow{2}{*}{$\olt$} & \multicolumn{2}{c|}{Easy} & \multicolumn{2}{c|}{Hard} & \multicolumn{2}{c}{Overall} \\
            \cmidrule{3-4} \cmidrule{5-6} \cmidrule{7-8}
            & & Acc@0.25 & Acc@0.50 & Acc@0.25 & Acc@0.50 & Acc@0.25 & Acc@0.50 \\
        \midrule
        \midrule
        SeeGround \cite{li2025seeground} & Mask3D \cite{schult2022mask3d} & 13.3 & 11.2 & 1.2 & 1.1 & 10.2 & 8.6 \\
        VLM-Grounder$^*$ \cite{xu2024vlmgrounder} & - & 14.5 & 12.1 & 6.1 & 2.3 & 12.3 & 9.6 \\
        VLM-Grounder$^*$ \cite{xu2024vlmgrounder} & Mask3D \cite{schult2022mask3d} & 16.4 & 12.0 & 10.2 & 4.8 & 14.8 & 10.1 \\
        SeqVLM \cite{lin2025seqvlm} & Mask3D \cite{schult2022mask3d} & 13.6 & 11.2 & 1.4 & 1.1 & 10.5 & 8.6 \\
        GPT4Scene \cite{qi2025gpt4scene} & Mask3D \cite{schult2022mask3d} & 10.1 & 7.7 & 0.9 & 0.7 & 7.7 & 5.9 \\
        SPAZER \cite{jin2025spazer} & Mask3D \cite{schult2022mask3d} & 15.3& 11.8& 2.3& 2.0& 12.0 & 9.3 \\
        Ours+SPAZER & Mask3D \cite{schult2022mask3d} & 49.8 & 36.3 & 29.5 & 20.8 & 44.5 & 32.3 \\
        Ours+SeeGround & Mask3D \cite{schult2022mask3d} & 46.5 & 34.5 & 27.5 & 20.1 & 41.6 & 30.8 \\
        Ours+SeqVLM & Mask3D \cite{schult2022mask3d} & 47.8 & 35.2 & 28.2 & 20.9 & 42.8 &  31.5 \\
        \textbf{Ours} & Mask3D \cite{schult2022mask3d} & \textbf{51.8} & \textbf{38.9} & \textbf{30.2} & \textbf{20.6} &\textbf{46.2} & \textbf{34.2} \\
        \midrule
        SeeGround \cite{li2025seeground} & GT & 20.2 & 19.8 & 11.3 & 10.4 & 17.9 & 17.4 \\
        VLM-Grounder$^*$ \cite{xu2024vlmgrounder} & GT & 31.4 & 21.3 & 20.6 & 17.8 & {28.6} & {20.4} \\
        SeqVLM \cite{lin2025seqvlm} & GT & 21.5 & 21.3 & 13.4 & 13.2 & 19.4 & 19.2 \\
        GPT4Scene \cite{qi2025gpt4scene} & GT & 13.6 & 13.4 & 7.9 & 7.3 & 12.1 & 11.8 \\
        SPAZER \cite{jin2025spazer} & GT & 46.3& 33.7& 18.7& 14.3& 39.2 & 28.7 \\
        Ours+SPAZER & GT & 51.8 & 51.3 & 40.9 & 40.8 & 49.0 & 48.6 \\
        Ours+SeeGround & GT & 51.1 & 50.7 & 38.6 & 38.2 & 47.9 & 47.5 \\
        Ours+SeqVLM & GT & 51.9 & 51.3 & 39.1 & 38.8 & 48.6 & 48.1 \\
        \textbf{Ours} & GT & \textbf{57.9} & \textbf{57.4} & \textbf{45.7} & \textbf{45.3} & \textbf{54.8} & \textbf{54.3} \\
        \bottomrule
    \end{tabular}
    }
    \label{tab:opentarget_comparison_add}
\end{table}

\section{Discussion of Limitations and Potential Solutions}

In the main paper, we discuss several limitations of OpenGround, including the static-scene assumption, the assumption that relevant objects are spatially proximal, and the dependence of Context-Guided Perception (CGP) on the underlying segmentation model.
These limitations do not affect the core formulation of our open-world grounding framework, but they may influence its robustness and applicability in more challenging real-world settings.
To better contextualize these limitations, we further discuss below several possible extensions and engineering directions that could alleviate them.
These discussions are not part of the main contribution of this paper, but may provide useful guidance for future work.

\paragraph{Static-Scene Assumption.}
OpenGround is currently designed for static scenes, which is consistent with the setting of existing 3DVG benchmarks.
To relax this assumption, one possible direction is to move from a single reconstructed scene to a 4D spatio-temporal representation, where RGB-D video, scene flow, or object tracking can support grounding within short locally static time windows.
Another possible direction is static--dynamic decomposition, where dynamic SLAM or motion segmentation separates stable background geometry from moving objects, allowing the framework to operate on the static component while handling dynamic objects through local updates.
A simpler alternative is a tracking-then-grounding paradigm, which first grounds the target in a near-static frame and then maintains object identity over time through 2D or 3D tracking.
These extensions are promising, but introduce substantial system complexity beyond the current scope of this work.

\paragraph{Spatial-Proximity Assumption.}
Our current design assumes that relevant objects are usually spatially close, which is common in existing 3DVG queries.
However, for long-range references involving objects that are spatially far apart, additional global reasoning may be required.
One possible solution is to introduce a bird's-eye-view (BEV) scene representation to provide coarse global layout information during reasoning.
Similarly, CGP could be extended with BEV-guided region proposal or spatial partitioning, so that the system first identifies plausible large-scale regions and then performs detailed local perception within them.
Such mechanisms may improve the handling of long-range dependencies while preserving the efficiency of progressive grounding.

\paragraph{Dependence on Segmentation Quality.}
The effectiveness of CGP depends on the quality of the segmentation model used to discover missing objects.
Although segmentation itself is not the focus of our work, this dependence can affect final grounding performance.
A practical direction to improve robustness is ensemble segmentation, where predictions from multiple segmentation models or model variants are combined to suppress model-specific errors and improve mask stability.
Another possible extension is to incorporate confidence estimation or verification modules, so that uncertain segmentation results can be re-checked before being inserted into the online $\olt$.
These strategies may improve the reliability of CGP without changing the overall grounding framework.

\paragraph{Error Recovery in Task Chain.}
Since later steps in the task chain depend on earlier grounded objects, an incorrect early prediction may propagate and affect subsequent reasoning.
A possible remedy is to introduce a lightweight backtracking mechanism.
For example, if the VLM determines during Single-Step Grounding that none of the current candidates satisfies the required conditions, the system can return to the previous step, discard the inconsistent hypothesis, and re-ground that object before proceeding.
This provides a simple form of error recovery for progressive grounding.
More generally, future work may study confidence-aware task chains or multi-hypothesis planning to further improve robustness against cascading errors.

\paragraph{Error Propagation and Recovery in Task Chains.}
Since later steps in a task chain depend on previously grounded objects, an incorrect intermediate prediction may propagate and affect subsequent grounding decisions. Nevertheless, our diagnostic results suggest that such cascading errors are present but are not the dominant source of failure. As shown in \cref{fig:task_chain_quantitative_acc}, performance does not degrade severely as the task-chain length increases. We further inspect 50 randomly sampled failed cases and 50 randomly sampled successful cases, and observe incorrect intermediate grounding in 32\% of the failed cases and 18\% of the successful cases. The higher occurrence in failed cases indicates that intermediate errors can increase the risk of final failure, whereas their presence in successful cases suggests that an imperfect intermediate result does not necessarily prevent the system from eventually identifying the target.

A lightweight backtracking mechanism could further improve robustness. For example, if the VLM determines during Single-Step Grounding that none of the current candidates satisfies the query and accumulated contextual constraints, the system can return to the most recent uncertain step, invalidate the inconsistent hypothesis, and re-ground the corresponding object before re-executing the subsequent subgoals. This provides a simple form of error recovery while retaining the progressive structure of task-chain grounding. More generally, confidence-aware task execution, bounded backtracking, or multi-hypothesis planning could be explored to reduce the effect of cascading errors in future work.

\section{More Visualization Results}

\begin{figure*}
    \centering
    \includegraphics[width=0.85\linewidth]{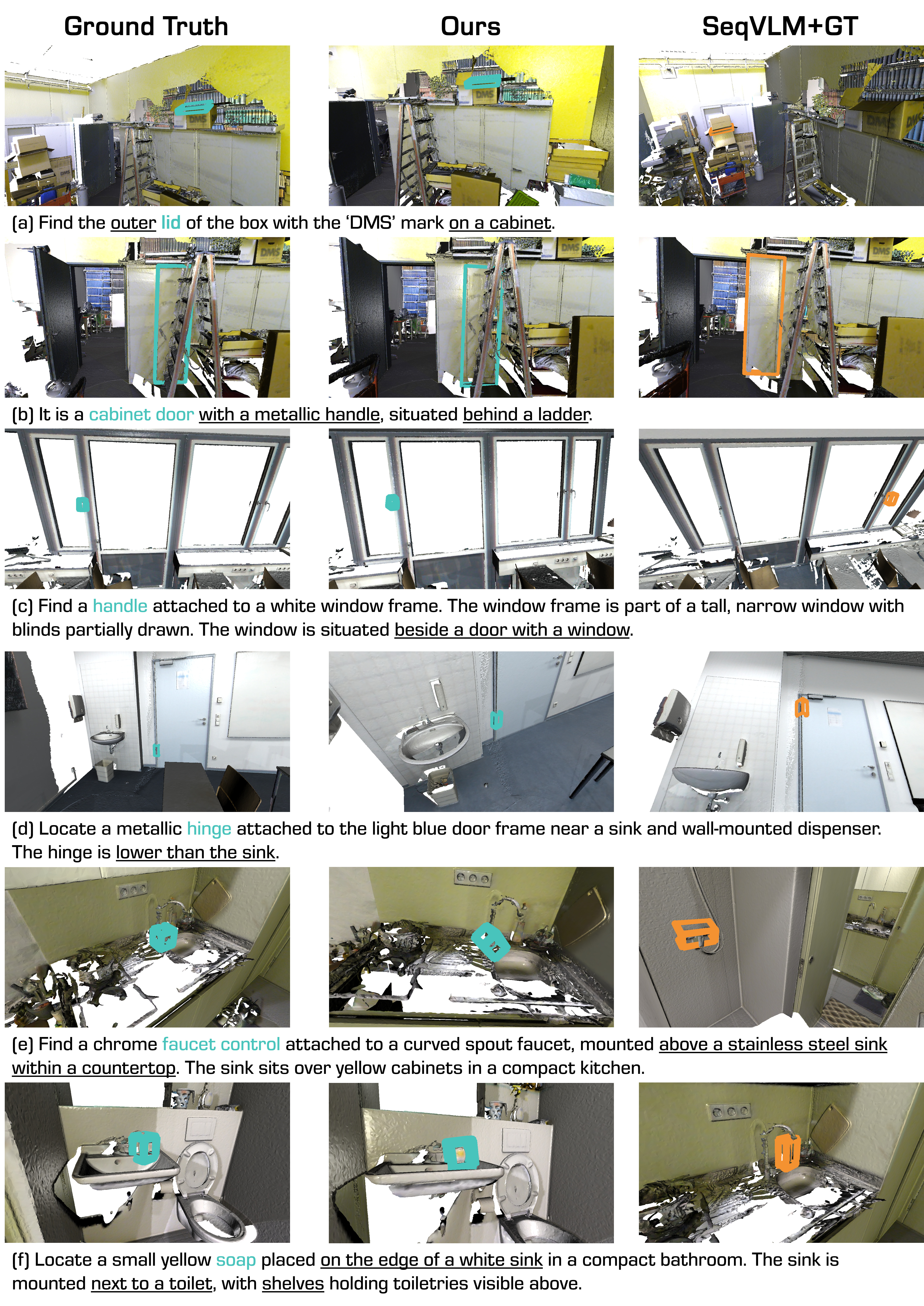}
    \caption{\textbf{Visualization of OpenGround on the OpenTarget benchmark.} Correct predictions are shown in teal and incorrect predictions in orange. Key linguistic cues used for grounding are underlined.}
    \label{fig:viz_w_gt}
\end{figure*}

\begin{figure*}
    \centering
    \includegraphics[width=0.85\linewidth]{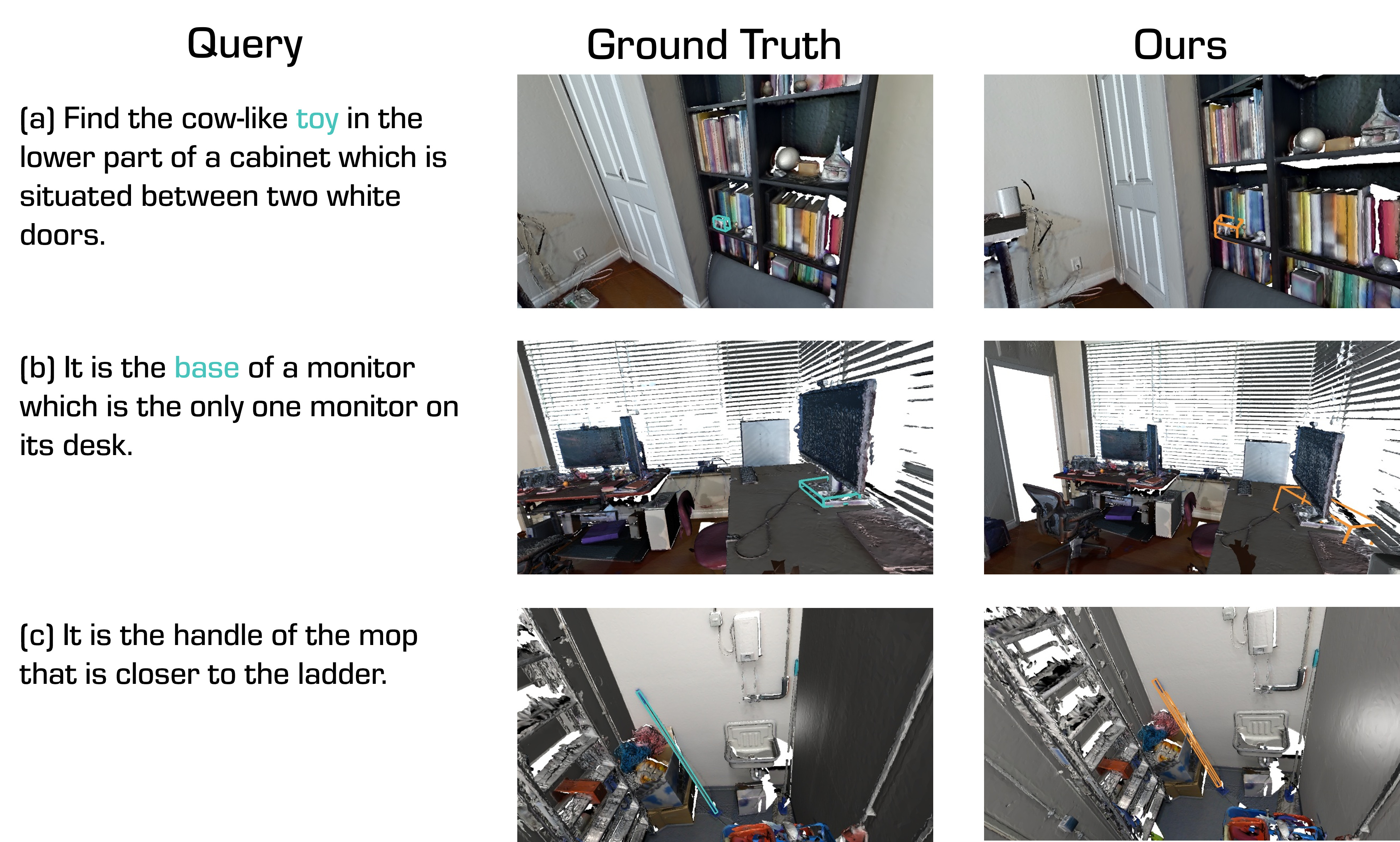}
    \caption{\textbf{Visualization of OpenGround on Application in ScanNet++ \cite{yeshwanthliu2023scannetpp}.} The target object (Ground Truth) is even out of OpenTarget, actually open-world object.}
    \label{fig:viz_wo_gt}
\end{figure*}

\begin{figure*}
    \centering
    \includegraphics[width=0.85\linewidth]{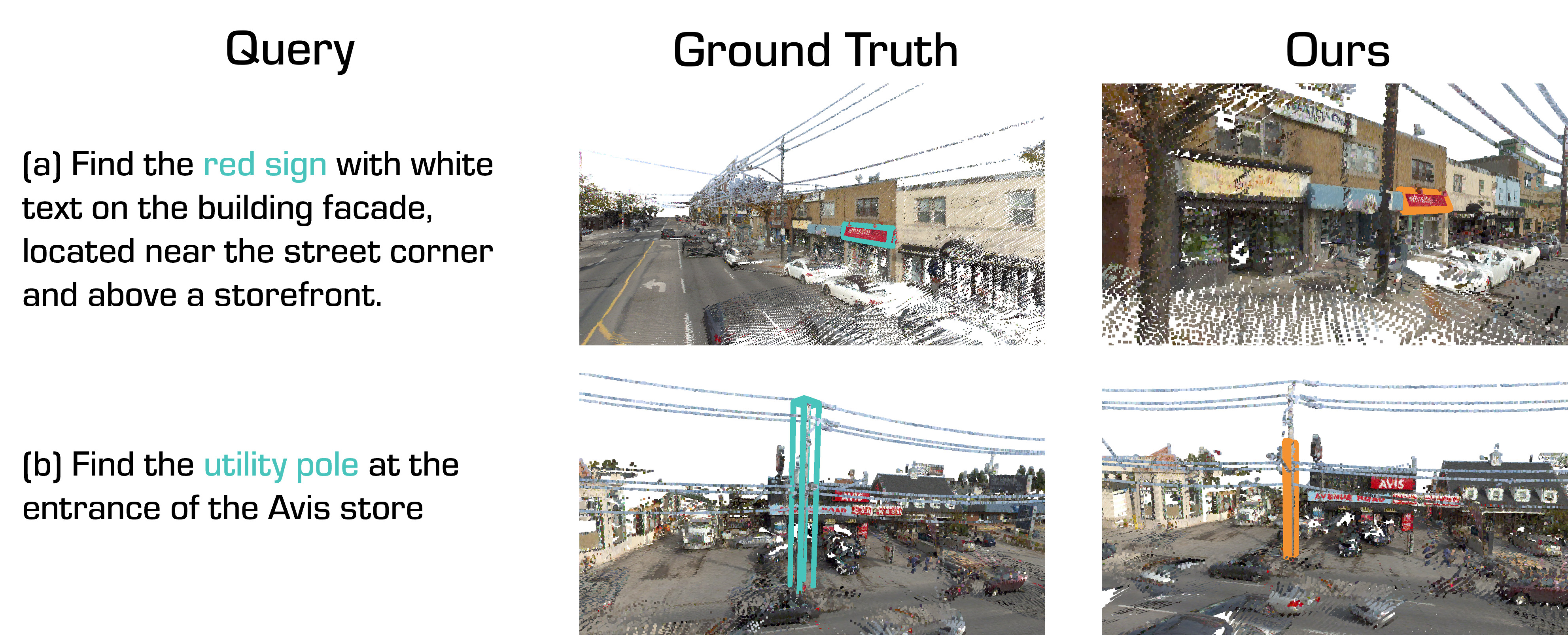}
    \caption{\textbf{Visualization of OpenGround on Application in Torondo3D \cite{tan2020toronto3d}.} The target object (Ground Truth) is also out of predefined categories in Toronto3D, actually open-world object.}
    \label{fig:viz_outdoors}
\end{figure*}

We provide additional qualitative results to further illustrate the behavior of OpenGround in both benchmark and real-world open-world settings.
\cref{fig:viz_w_gt} presents comparisons on the OpenTarget benchmark, where we show ground-truth views, our predictions, and SeqVLM+GT. Correct and incorrect predictions are highlighted, and key linguistic cues are underlined to reveal how the model reasons over queries. Beyond benchmark evaluation, \cref{fig:viz_wo_gt} demonstrates OpenGround applied to ScanNet++, where target objects even fall outside OpenTarget categories. These examples highlight OpenGround’s ability to generalize to truly open-world objects and perform reliable localization even in cluttered, unseen environments. In addition, we further evaluate our method in fully open-world outdoor scenes where non $\olt$ provided, as shown in \cref{fig:viz_outdoors}.

\newpage

\lstset{
  basicstyle=\ttfamily\small,
  breaklines=true,
  breakatwhitespace=false,
  columns=fullflexible
}
\begin{center}
\captionof{table}{Prompt for dataset annotation. ``\{images\}'' represents the input images, ``\{label hierarchy\}'' is the label hierarchy of the target object, and ``\{context reference\}'' is the generated annotation of the parent of the target object.}
\label{tab:prompt_annotation}
\end{center}
\begin{tcolorbox}[
    colback=boxbg,        
    colframe=black,       
    boxrule=0.6pt,        
    arc=3pt,              
    enhanced,
    title=Prompt Template for Dataset Annotation,
    coltitle=headerfg,
    colbacktitle=headerbg,
    fonttitle=\bfseries,
    left=4pt,right=4pt,top=4pt,bottom=4pt
]
Please analyze user's input images, label hierarchy and context reference. Label hierarchy consists of current object label and its parent objects' labels. Context reference is the description of current object's direct parent. $\langle$image\_1$\rangle$ is the image observing the target object, and $\langle$image\_2$\rangle$...$\langle$image\_N$\rangle$ are the images observing distractors with same label. All objects in images are marked with red bounding boxes. 

Your task is generate a discriminative description for the object in the bounding box of $\langle$image\_1$\rangle$. The description should distinguish the object from distractors and keep information from context reference.

Attention:

    1. Words 'bounding box' and 'image' should not appear in your answer.
    
    2. It would be better to figure out the target object explicitly as example does.
    
    3. It would be better to vary sentence structure, compared to the context reference and the examples.

\textbf{Here are examples:}

Example 1:

Images: $\langle$image\_1$\rangle$, $\langle$image\_2$\rangle$, $\langle$image\_3$\rangle$, $\langle$image\_4$\rangle$

Label hierarchy: "cabinet"$\rightarrow$"drawer"$\rightarrow$"handle". 

Context reference: "Find a wooden drawer which is the second from top to down of a cabinet. The cabinet nears a backpack."

Output: Locate a rectangular handle is attached to the second wooden drawer from top to down of a cabinet. The cabinet nears a backpack.

Example 2:

Images: $\langle$image\_1$\rangle$, $\langle$image\_2$\rangle$

Label hierarchy: "door". 

Context reference: None

Output: Locate a wooden door which is near a whiteboard.

...

\textbf{Now, start your task:}

Images: \{images\}. Label hierarchy: \{label hierarchy\}. Context reference: \{context reference\}.
\end{tcolorbox}

\begin{center}
\captionof{table}{Prompt for dataset verification. ``\{images\}'' represents the input images and ``\{query\}'' is the generated annotation of the target object.}
\label{tab:prompt_verification}
\end{center}
\begin{tcolorbox}[
    breakable,
    colback=boxbg,
    colframe=black,
    boxrule=0.6pt,
    arc=3pt,
    enhanced,
    title=Prompt Template for Dataset Verification,
    coltitle=headerfg,
    colbacktitle=headerbg,
    fonttitle=\bfseries,
    left=4pt,right=4pt,top=4pt,bottom=4pt
]
Please analyze user's input images and query. Your task is reasoning on query and determine the image whose object in the red bounding box satisfies the query. Return the result strictly in this JSON format without any addition content:
\begin{lstlisting}
{
    "answer": {
        "reason": "xxx",
        "image": xxx
    }
}
\end{lstlisting}
\textbf{Here are example:}

Example 1:

Images: $\langle$image\_1$\rangle$, $\langle$image\_2$\rangle$, $\langle$image\_3$\rangle$, $\langle$image\_4$\rangle$

Query: Locate a rectangular handle is attached to the second wooden drawer from top to down of a cabinet. The cabinet nears a backpack.

Response:
\begin{lstlisting}
{
    "answer": {
        "reason": "<image_1> ... <image_4> are handles on the same cabinet. 
        <image_1> is the second from top, <image_2> is the third from top, 
        <image_4> is the forth from top, and <image_4> is the first top one. 
        So the answer is <image_1>",
        "image": 1
    }
}
\end{lstlisting}

...

\textbf{Now, start your task:}

Image: \{images\}. Query: \{query\}.

\end{tcolorbox}

\begin{center}
\captionof{table}{Prompt for parsing relevant and target objects from the query description. ``\{query\}'' represents the input query $\query$.}
\label{tab:prompt_object_parsing}
\end{center}
\begin{tcolorbox}[
    breakable,
    colback=boxbg,        
    colframe=black,       
    boxrule=0.6pt,        
    arc=3pt,              
    enhanced,
    title=Prompt Template for Objects Parsing,
    coltitle=headerfg,
    colbacktitle=headerbg,
    fonttitle=\bfseries,
    left=4pt,right=4pt,top=4pt,bottom=4pt
]

Please analyze the user's input query and identify all specific physical objects mentioned in it (exclude abstract concepts, actions, areas, or standalone attributes). Explicitly relevant objects should also be considered. Return the result strictly in this JSON format without any additional content:
\begin{lstlisting}
{
    "objects": [
        {"name":"object1", "is_target": true/false}, 
        {"name":"object2", "is_target": true/false}
    ]
}
\end{lstlisting}

Only one object may be marked as target (is\_target=true). All other objects must be marked as non-target (is\_target=false).

\textbf{Here are examples}:

Example 1:

Query: "Locate the box which is the top one of the stack of boxes on the floor. Additionally, the box is near a black door and a sink in a storage room."

Response:
\begin{lstlisting}
{
    "objects": [
        {"name":"box", "is_target": true}, 
        {"name":"boxes stack", "is_target": false},
        {"name":"door", "is_target": false},
        {"name":"sink", "is_target": false}
    ]
}
\end{lstlisting}

Example 2:

Query: "A small metallic hinge attached to the door frame at the lower part"

Response:
\begin{lstlisting}
{
    "objects": [
        {"name":"hinge", "is_target": true}, 
        {"name":"door frame", "is_target": false},
        {"name":"door", "is_target": false} % relevant to door frame
    ]
}
\end{lstlisting}

...

\textbf{Now, start your task:}

Query: \{query\}.

\end{tcolorbox}

\begin{center}
\captionof{table}{Prompt for task-chain construction. ``\{query\}'' denotes the input query $\query$, ``\{context objects\}'' denotes the extracted context labels with their candidate counts, and ``\{target object\}'' denotes the target label with its candidate count.}
\label{tab:prompt_task_chain_construction}
\end{center}

\begin{tcolorbox}[
    breakable,
    colback=boxbg,
    colframe=black,
    boxrule=0.6pt,
    arc=3pt,
    enhanced,
    title=Prompt Template for Task-Chain Construction,
    coltitle=headerfg,
    colbacktitle=headerbg,
    fonttitle=\bfseries,
    left=4pt,right=4pt,top=4pt,bottom=4pt
]
You are given a natural-language grounding query, a target object, and several context objects.
Each object is associated with a candidate count retrieved from an offline object lookup table (OLT).

Your task is to construct a \textbf{human-like grounding plan} as an ordered sequence of object-level sub-goals for 3D visual grounding.

You must reason over the following factors:

\begin{enumerate}
    \item \textbf{Semantic Dependency}: 
    Objects linked by spatial, possessive, or compositional relations should usually be grounded in dependency order.
    For example, in ``handle of the drawer'', the drawer should usually be grounded before the handle.
    
    \item \textbf{Candidate Availability}: 
    If an object has \textbf{0 candidates}, it is currently missing from the offline OLT and will require later perception guided by already grounded context.
    Therefore, such objects should usually be placed \textbf{after} the objects they semantically depend on.

    \item \textbf{Grounding Difficulty}: 
    A larger candidate count implies a broader search space and higher ambiguity.
    More ambiguous objects should usually be grounded later, after easier or more distinctive context objects have narrowed the search space.
\end{enumerate}

Additional instructions:

\begin{itemize}
    \item Prefer a plan that resembles how humans would progressively localize the target using contextual anchors.
    \item The \textbf{target object should preferably appear last}, because the goal is usually resolved after sufficient context has been established.
    \item However, this is only a \textbf{soft constraint}: if the target has very strong uniqueness (e.g., much fewer candidates than other objects, or is clearly more distinctive), it may be placed earlier.
    \item Do not output duplicated objects.
    \item Keep the sequence complete: include all provided context objects and the target object exactly once.
\end{itemize}

Return the result \textbf{strictly} in the following JSON format, with no additional text:

\begin{lstlisting}
{
  "reason": "Briefly explain the ordering based on semantic dependency, candidate availability, and grounding difficulty.",
  "sequence": [
    {"name": "object_name_1", "origin_index": 0},
    {"name": "object_name_2", "origin_index": 2},
    {"name": "target_object", "origin_index": 1}
  ]
}
\end{lstlisting}

Here, \texttt{origin\_index} is the index of the object in the provided context-object list, and use \texttt{-1} for the target object.

\textbf{Example 1}

Query: ``Locate the box which is the top one of the stack of boxes on the floor. Additionally, the box is near a black door and a sink in a storage room.''

Context Objects: [("boxes stack", 0), ("door", 3), ("sink", 2)]

Target Object: ("box", 17)

Response:
\begin{lstlisting}
{
  "reason": "The sink and door are easier contextual anchors with small candidate sets. The boxes stack has no candidate and should be grounded after its supporting context. The target box has the largest candidate set and depends on the stack, so it is placed last.",
  "sequence": [
    {"name": "sink", "origin_index": 2},
    {"name": "door", "origin_index": 1},
    {"name": "boxes stack", "origin_index": 0},
    {"name": "box", "origin_index": 3}
  ]
}
\end{lstlisting}

\textbf{Example 2}

Query: ``Locate a slender wooden handle attached to a white drawer in a kitchen cabinet unit beneath a smooth countertop and beside a black oven. The handle is situated between other two handles. The cabinet rests against a dark backsplash, with a wall-mounted rack nearby.''

Context Objects: [("oven", 1), ("countertop", 1), ("cabinet", 10), ("rack", 0), ("backsplash", 0), ("drawer", 0)]

Target Object: ("handle", 0)

Response:
\begin{lstlisting}
{
  "reason": "The oven and countertop are distinctive anchors with minimal ambiguity. The backsplash and rack have no candidates, so they should be handled after basic anchors are established. The cabinet has many candidates and should be delayed until context narrows the search space. The drawer depends on the cabinet, and the handle depends on the drawer, so the target is placed last.",
  "sequence": [
    {"name": "oven", "origin_index": 0},
    {"name": "countertop", "origin_index": 1},
    {"name": "backsplash", "origin_index": 4},
    {"name": "rack", "origin_index": 3},
    {"name": "cabinet", "origin_index": 2},
    {"name": "drawer", "origin_index": 5},
    {"name": "handle", "origin_index": 6}
  ]
}
\end{lstlisting}

\textbf{Now solve the following case.}

Query: \{query\}

Context Objects: \{context objects\}

Target Object: \{target object\}
\end{tcolorbox}

\begin{center}
\captionof{table}{Prompt for conditions retrieval. ``\{query\}'' represents the input query $\query$, ``\{related objects\}'' represents the objects in the task chain, and ``\{target\}'' represents the object to retrieve conditions in this step.}
\label{tab:prompt_conditions_retrieval}
\end{center}
\begin{tcolorbox}[
    breakable,
    colback=boxbg,        
    colframe=black,       
    boxrule=0.6pt,        
    arc=3pt,              
    enhanced,
    title=Prompt Template for Conditions Retrieval,
    coltitle=headerfg,
    colbacktitle=headerbg,
    fonttitle=\bfseries,
    left=4pt,right=4pt,top=4pt,bottom=4pt
]
Please analyze the user's input query, related objects and the target object. Your task is summarize the conditions of the target object. Return the result strictly in the followng JSON format without any additional content:
\begin{lstlisting}
{
    "conditions": [
        "condition1", "condition2", ...
    ]
}
\end{lstlisting}
The output conditions should be detailed and sufficient.

\textbf{Here are examples:}

Example 1:

Query: "Locate the box which is the top one of the stack of boxes on the floor. Additionally, the box is near a black door and a sink in a storage room."

Related Objects: ["sink", ..., "box"]

Target Object: "boxes stack"

Response:
\begin{lstlisting}
{
    "conditions": [
        "it is a stack of boxes",
        "it is on the floor",
        "it is near a black door",
        "it is near a sink"
    ]
}
\end{lstlisting}

...

\textbf{Now, start your task:}

Query: \{query\}. Related Objects: \{related objects\}. Target Object: \{target\}.

\end{tcolorbox}

\begin{center}
    
\captionof{table}{Prompt for VLM reasoning. ``\{images\}'' represents the combined image set with annotate images $\rgbset^\star_A$ and non-annotated images $\rgbset^\star$, ``\{query\}'' represents the input query $\query$, and ``\{conditions\}'' represents the conditions obtained in \cref{tab:prompt_conditions_retrieval}.}
\label{tab:prompt_vlm_reasoning}
\end{center}
\begin{tcolorbox}[
    breakable,
    colback=boxbg,        
    colframe=black,       
    boxrule=0.6pt,        
    arc=3pt,              
    enhanced,
    title=Prompt Template for VLM Reasoning,
    coltitle=headerfg,
    colbacktitle=headerbg,
    fonttitle=\bfseries,
    left=4pt,right=4pt,top=4pt,bottom=4pt
]

Please analyze user's input query, the conditions of the target object and images. For each N, $\langle$image\_2N-1$\rangle$ is origin image without any annotation, $\langle$image\_2N$\rangle$ is the annotated images with previously grounded objects (which might be appeared in the conditions) and current candidates. Each annotation is consists of a bounding box, a label and corresponding id. Your task is matching each candidate with the conditions. Return the result strictly in the following JSON format without any additional content:
\begin{lstlisting}
{
    "id1": ["condition1", "condition3"],
    "id2": ["condition2"],
    ...
}
\end{lstlisting}
You should observe and compare the images carefully. The output conditions should be exactly the same as the input.

\textbf{Here are examples:}

Example 1:

Images: $\langle$image\_1$\rangle$, $\langle$image\_2$\rangle$, $\langle$image\_3$\rangle$, $\langle$image\_4$\rangle$

Query: "Locate the box which is the top one of the stack of boxes on the floor. Additionally, the box is near a black door and a sink in a storage room."

Conditions: [
    "it is a stack of boxes",
    "it is on the floor",
    "it is near a black door",
    "it is near a sink"
]

Response: 
\begin{lstlisting}
{
    "128": ["it is a stack of boxes", "it is on the floor"],
    "127": ["it is a stack of boxes"],
    ...
    "131": [
        "it is a stack of boxes",
        "it is on the floor",
        "it is near a black door",
        "it is near a sink"
    ]
}
\end{lstlisting}
...

\textbf{Now, start your task:}

Images: \{images\}. Query: \{query\}. Conditions: \{conditions\}.

\end{tcolorbox}
\newpage

\end{document}